\documentclass[sigconf, nonacm]{acmart}

\usepackage{algpseudocode}
\usepackage[ruled,linesnumbered]{algorithm2e}
\newtheorem{exmp}{Example}
\algnewcommand\algorithmicinput{\textbf{Input:}}
\algnewcommand\Input{\item[\algorithmicinput]}
\algnewcommand\algorithmicoutput{\textbf{Output:}}
\algnewcommand\Output{\item[\algorithmicoutput]}
\algnewcommand\algorithmicforeach{\textbf{for each}}
\algdef{S}[FOR]{ForEach}[1]{\algorithmicforeach\ #1\ \algorithmicdo}

\usepackage{xcolor}

\newcommand\vldbdoi{XX.XX/XXX.XX}
\newcommand\vldbpages{XXX-XXX}
\newcommand\vldbvolume{14}
\newcommand\vldbissue{1}
\newcommand\vldbyear{2020}
\newcommand\vldbauthors{\authors}
\newcommand\vldbtitle{\shorttitle} 
\newcommand\vldbavailabilityurl{URL_TO_YOUR_ARTIFACTS}
\newcommand\vldbpagestyle{plain} 

\begin{document}
\title{FOLD-SE: An Efficient Rule-based Machine Learning Algorithm \\
with Scalable Explainability}

\author{Huaduo Wang}
\affiliation{%
  \institution{The University of Texas at Dallas}
  \streetaddress{800 W. Campbell Road}
  \city{Richardson}
  \country{USA}}
\email{huaduo.wang@utdallas.edu}

\author{Gopal Gupta}
\affiliation{%
  \institution{The University of Texas at Dallas}
  \streetaddress{800 W. Campbell Road}
  \city{Richardson}
  \country{USA}}
\email{gupta@utdallas.edu}

\begin{abstract}
We present FOLD-SE, an efficient, explainable machine learning algorithm for classification tasks given tabular  data containing numerical and categorical values. FOLD-SE generates a set of default rules---essentially a stratified normal logic program---as an (explainable) trained model.  Explainability provided by FOLD-SE is scalable, meaning that regardless of the size of the dataset, the number of learned rules and learned literals stay quite small while good accuracy in classification is maintained. A model with smaller number of rules and literals is easier to understand for human beings. FOLD-SE is competitive with state-of-the-art machine learning algorithms such as XGBoost and Multi-Layer Perceptrons (MLP) wrt accuracy of prediction. However, unlike XGBoost and MLP, the FOLD-SE algorithm is explainable. The FOLD-SE algorithm builds upon our earlier work on developing the explainable FOLD-R++ machine learning algorithm for binary classification and inherits all of its positive features. Thus, pre-processing of the dataset, using techniques such as one-hot encoding, is not needed. 
Like FOLD-R++, FOLD-SE uses prefix sum to speed up computations resulting in FOLD-SE being an order of magnitude faster than XGBoost and MLP in execution speed. 
The FOLD-SE algorithm outperforms FOLD-R++ as well as other rule-learning algorithms such as RIPPER in efficiency, performance and scalability, especially for large datasets. A major reason for scalable explainability of FOLD-SE is the use of a literal selection heuristics based on \textit{Gini Impurity}, as opposed to \textit{Information Gain} used in FOLD-R++. A multi-category classification version of FOLD-SE is also presented. 
\end{abstract}

\maketitle
\section{Introduction}

Success of machine learning has led to a plethora of Artificial Intelligence (AI) applications. However, the effectiveness of these systems is limited by the machines' current inability to explain their decisions and actions to human users. That is mainly because the statistical machine learning methods produce models that are complex algebraic solutions to optimization problems such as risk minimization or data likelihood maximization. Lack of intuitive descriptions makes it hard for users to understand and verify the underlying rules that govern the model. Also, these methods cannot produce a justification for a prediction they compute for a new data sample.


Rule-based machine learning (RBML) algorithms have been devised that \textit{learn} a set of relational rules that collectively represent the logic of the concept encapsulated in the data. 
The generated rules are more comprehensible to humans compared to the complicated deep learning models or complicated formulas. Examples of such algorithms include FOIL \cite{foil} and RIPPER \cite{RIPPER}. Some of these algorithms allow the knowledge learned to be incrementally extended without retraining the entire model. The learned symbolic rules make it easier for users to understand and verify them. Inductive Logic Programming algorithms \cite{ilp20,ilpsurvey22} are also RBML algorithms.

The RBML problem can be regarded as a search problem where we search for a set of rules consistent with the training examples. Usually, the search for rules can be processed either top-down or bottom-up. A bottom-up approach starts by creating most-specific rules from training examples and exploring the hypothesis space by employing generalization techniques. Bottom-up approaches do not perform very well on large datasets.
A top-down approach starts by building the most-general rules from training examples and then specializes them into a final rule set. Most of the RBML algorithms are not efficient for large datasets even if they follow a top-down approach. Some of them, e.g., TREPAN \cite{trepan} 
extract rules from statistical machine learning models, but their performance and efficiency is limited by the target machine learning model. Yet other algorithms train models using a logic programming-based solver, for example, most of the Inductive Logic Programming (ILP) based algorithms fall in this class \cite{ilp20,ilpsurvey22}.

An important aspect of RBML systems is that they should be scalable, meaning that they should work with large datasets, and learn the rules in a reasonable amount of time.  For RBML algorithms, the size of the generated rule-set---represented by the number of rules and number of condition predicates (features) involved in the rules---has a big impact on human-understanding of the rules (interpretability) and explaining predictions (explainability). The more rules and predicates a rule-set representing a model contains, the harder it is for a human to understand. Generally, for most RBML systems, as the size of dataset increases, the number of rules and conditional predicate increases. Ideally, we would like for the rule-set size to not increase with dataset size. We call this concept \textit{scalable explainability}, i.e., the size of the rule-set is a small constant regardless of the dataset size. Thus, even when size of the input training data is very large, the rule-set representing the model should be small enough for a human to comprehend it.

The FOIL algorithm by Quinlan \cite{foil} is a popular top-down RBML algorithm. FOIL uses heuristics from information theory called \textit{weighted information gain}. The use of greedy heuristics allows FOIL to run much faster than bottom-up approaches and scale up much better. The FOLD RBML algorithm by Shakerin, Salazar, and Gupta \cite{fold}---the foundation of the FOLD-SE algorithm reported in this paper---is inspired by the FOIL algorithm. The FOLD algorithm learns a default theory with exceptions, represented as a \textit{stratified normal logic program}. The FOLD algorithm incrementally generates literals for \textit{default rules} that cover positive examples while avoiding covering negative examples. It then swaps the positive and negative examples and calls itself recursively to learn exceptions to the default 
when there are still negative examples falsely covered.
The process is applied recursively to learn exceptions to exceptions, exceptions to exceptions to exceptions, and so on. We subsequently developed the FOLD-R++ RBML algorithm on top of the FOLD algorithm. The FOLD-R++ algorithm utilizes the prefix sum computation technique with special comparison operators to speed up literal selection while avoiding one-hot encoding for mixed-type data. It also introduces a hyper-parameter $ratio$ to speed up training while reducing the number of generated rules \cite{foldrpp,foldrm}. The FOLD-R++ algorithm provides dramatic improvement over the FOLD algorithm.

The FOLD-R++ RBML algorithm also generates far fewer rules than the well-known rule-based ML algorithm RIPPER while outperforming it in classification accuracy. For very large datasets, most of the RBML algorithms are not scalable. They fail to finish training in reasonable time. %
Even RIPPER and our FOLD-R++ algorithms often generate too many rules making them incomprehensible to humans. For example, Rain in Australia is a large dataset with over 140K training examples. With the same target class ‘No’, RIPPER generates 180 rules with over 700 literals that achieve 63\% accuracy while FOLD-R++ generates 48 rules with around 120 literals that achieve 79\% accuracy. That's too many rules, arguably, for a human to understand. Another explainability-related problem with these RBML algorithms is that the rules they generate change significantly when a small percentage of training data changes. Thus, the learned rule-set will be different for different splits of the dataset for training and testing. We would like for the rule-set to not change as the training dataset changes. That is, if the dataset does not change in a materially significant manner, then the rule-set should not change. 

To deal with the above explainability issue on large datasets, this paper presents the FOLD-SE algorithm that employs a newly created heuristic (based on Gini Impurity) for literal selection that greatly reduces the number of rules and predicates in the generated rule-set. In addition, the FOLD-SE algorithm also improves explainability by introducing a rule pruning mechanism in the training process. The pruning mechanism ameliorates the long-tail effect, namely, that rules generated later in the learning process cover fewer examples than those generated earlier. As a result of this pruning, the rule-set generated for different training-testing split of the dataset remains largely unchanged from one split to another. Finally, we add two comparison operators that improve the literal selection process for sparse datasets containing many missing values.  The FOLD-SE algorithm provides scalable explainability. Thus:
\begin{enumerate} 
\item FOLD-SE generates a rule-set that uses a small number of rules and predicates (features) regardless of the dataset size. 
\item The generated rule-set is almost the same regardless of the training/testing split used.
\end{enumerate} 

\noindent It should be noted that a smaller rule set is easier for humans to understand and interpret. A smaller rules set will also make providing an explanation for a given prediction easier. We also appeal here to the principle known as Occam's razor, which stipulates that ``with competing theories or explanations, the simpler one, for example a model with fewer parameters, is to be preferred" \cite{occam}.

Our experimental results indicate that FOLD-SE is competitive in accuracy with state-of-the-art machine learning algorithms such as the XGBoost classifier and Multi-Layer Perceptron (MLP). It is an order of magnitude faster than both these methods in execution efficiency. In addition, the FOLD-SE algorithm provides justification (explanation) for a prediction. 
The FOLD-SE algorithm significantly outperforms our previously-developed FOLD-R++ algorithm, on which it is based, as well as the RIPPER algorithm. 

The main contribution of the paper is the FOLD-SE algorithm that learns a model described as a set of default rules. The FOLD-SE algorithm has the following salient features.

\vspace{-0.009in} 
\begin{itemize}
    \item Minimal work is needed to prepare the dataset for the FOLD-SE algorithm; Complex transformations such as one-hot encoding are not needed. The algorithm works with mixed data, where a feature can have both categorical and numerical values.
    \item Missing values and null values are easily handled; they are treated as categorical values automatically and do not require any extra care.
    \item FOLD-SE is competitive wrt accuracy and F1-score metrics to state-of-the-art machine learning algorithms such as XGBoost. It significantly outperforms other rule-based methods such as RIPPER in this aspect.
    \item Regardless of the dataset size, the number of rules and literals in the generated rule-set is small. The number of rules and literals does not grow as the dataset size increases. 
    \item FOLD-SE works for even very large datasets. Most RBML algorithms do not scale up to very large datasets as they run out of resources.
    \item The rule-set generated is quite stable in that different splits of the dataset into training and testing sets do not drastically change the rule-set.
    \item The implementation of the algorithms is quite efficient. Thanks to the use of prefix sum computations, FOLD-SE is an order of magnitude faster than state-of-the-art systems such as XGBoost. 
\end{itemize}

\noindent To the best of our knowledge, there is no other explainable machine learning algorithm for tabular data that has all these features. 

\section{Default Rules}

\subsection{Default Logic} 
The FOLD, FOLD-R++, and FOLD-SE algorithms represent the rule-set that they learn from a dataset as a \textit{default theory} that is expressed as a normal logic program, i.e., logic programming with negation-as-failure \cite{lloyd,gelfondkahl}. The logic program is stratified in that there are no recursive calls in the rules. We briefly describe default logic below. 
Note that in this paper, we use the terms literal, predicate and feature interchangeably. We assume that the reader is familiar with logic programming \cite{lloyd,gelfondkahl} and classification problems \cite{bishop2006}.

Default Logic \cite{reiter80} is a non-monotonic logic to formalize commonsense reasoning. A default $D$ is an expression of the form 
$$ A: \textbf{M} B \over\Gamma$$
\noindent which states that the conclusion $\Gamma$ can be inferred if pre-requisite $A$ holds and $B$ is justified. $\textbf{M} B$ stands for ``it is consistent to believe $B$".
Normal logic programs can encode a default theory quite elegantly \cite{gelfondkahl}. A default of the form: 
$$\alpha_1 \land \alpha_2\land\dots\land\alpha_n: \textbf{M} \lnot \beta_1, \textbf{M} \lnot\beta_2\dots\textbf{M}\lnot\beta_m\over \gamma$$
\noindent can be formalized as the
normal logic programming rule:
$$\gamma ~\texttt{:-}~ \alpha_1, \alpha_2, \dots, \alpha_n, \texttt{not}~ \beta_1, \texttt{not}~ \beta_2, \dots, \texttt{not}~ \beta_m.$$
\noindent where $\alpha$'s and $\beta$'s are positive predicates and \texttt{not} represents negation-as-failure. We call such rules \emph{default rules}. 
Thus, the default 

$$bird(X): M \lnot penguin(X)\over flies(X)$$

\noindent will be represented as the following default rule in normal logic programming:

{\tt flies(X) :- bird(X), not penguin(X).}

\noindent We call {\tt bird(X)}, the condition that allows us to jump to the default conclusion that {\tt X} flies, the {\it default part} of the rule, and {\tt not penguin(X)} the \textit{exception part} of the rule. 

\subsection{Default Rules as Machine Learning Models} 

Default rules allow knowledge to be modeled in an \textit{elaboration tolerant} manner \cite{baral,gelfondkahl}. Much of human commonsense knowledge learned over time is represented as default rules. Default rules are an excellent vehicle for representing inductive generalizations. Humans indeed represent inductive generalizations as default rules. Arguably, the sophistication of human thought process is in large part due to copious use of default rules by humans \cite{gelfondkahl}. 

\begin{figure}[h]
    \centering
    \includegraphics[width=8cm]{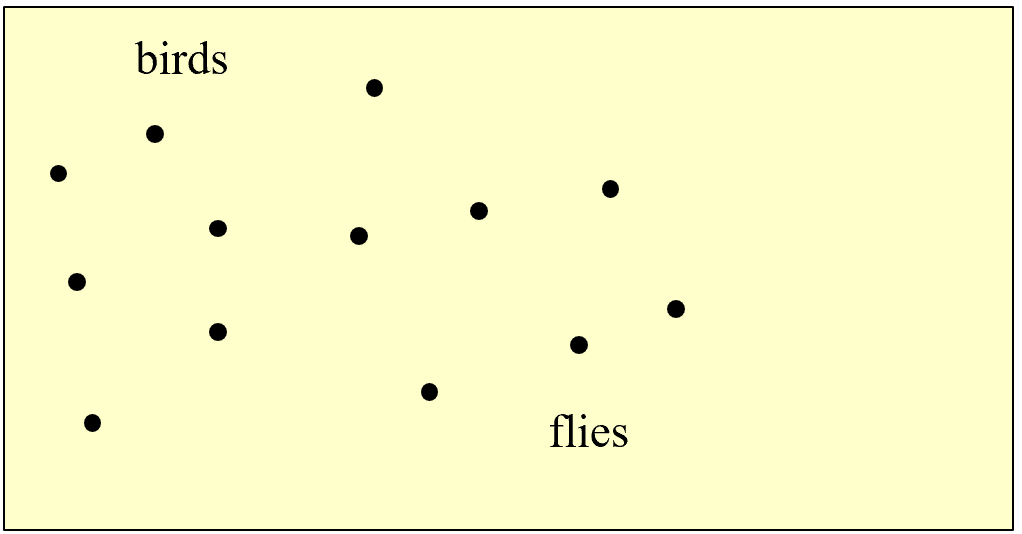}
    \caption{All Birds Fly}
    \label{fig1}
    \end{figure}

\begin{figure}[h]
    \centering
    \includegraphics[width=8cm]{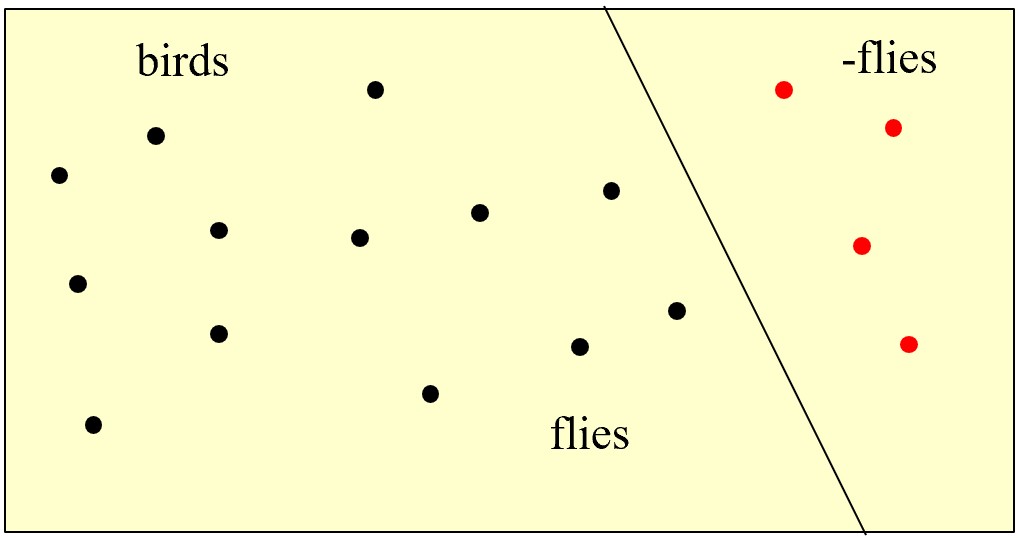}
    \caption{Some birds are penguins and don't fly; the decision boundary separates concept of flies from -flies where -flies represents the concept of not being able to fly; red dots are penguins}
    \label{fig2}
    \end{figure}

Consider the birds example above. We observe that bird 1 flies, bird 2 flies, bird 3 flies, and so on. From this we can generalize and learn the default rule that ``birds fly." But then we notice that a few of the birds that are penguins, do not fly. So we add an exception to our rule: ``birds fly, unless they are penguins.". What we are doing is really adjusting our decision boundary, as illustrated in Fig. \ref{fig1} and Fig. \ref{fig2}. In logic programming, we can make the exception part of the rule explicit, and code it as:

~~~~~~~~~~{\tt flies(X) :- bird(X), not abnormal\_bird(X).}

~~~~~~~~~~{\tt abnormal\_bird(X) :- penguin(X).}

\noindent Suppose, we later discover that there is a subclass of penguins (called super-penguins) that can fly. In such a case, we have learned an exception to an exception. This will be coded in logic programming as:

~~~~~~~~~~{\tt flies(X) :- bird(X), not abnormal\_bird(X).}

~~~~~~~~~~{\tt abnormal\_bird(X) :- penguin(X), not abnormal\_penguin(X).}

~~~~~~~~~~{\tt abnormal\_penguin(X) :- superpenguin(X).}

\begin{figure}[h]
    \centering
    \includegraphics[width=8cm]{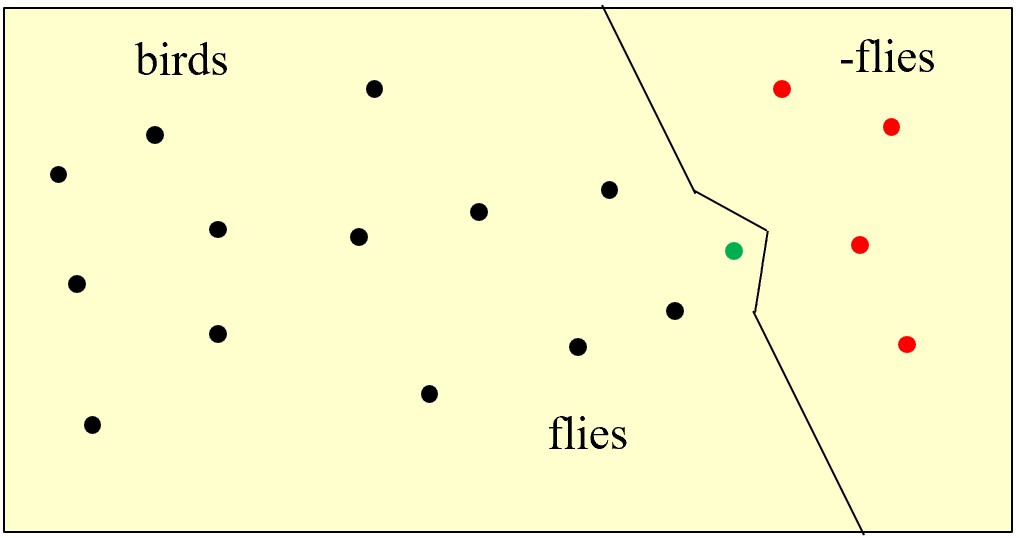}
    \caption{Super-penguins (green dots) are exceptional penguins that can fly; decision boundary is changed to refine the concepts of flies and -flies}
    \label{fig3}
    \end{figure}

\noindent Thus, default rules with exceptions, exceptions to exceptions, exceptions to exceptions to exceptions, and so on, allow us to dynamically refine our decision boundary as our knowledge of a concept evolves (See Fig. \ref{fig3}). This is the insight that the FOLD family of algorithms uses to learn a model underlying a dataset. An additional advantage of using default rules to describe a model is that since humans themselves use default rules for inductive generalizations, they can understand the model more readily. Figure \ref{fig1}-\ref{fig3} illustrate the idea. 

\begin{table*}[h!]
\centering
\setlength{\tabcolsep}{1pt}
\begin{tabular}{|l|c|c||c|c|c|c||c|c|c|c||c|c|c|c||c|c|c|c|}
\hline
\multicolumn{3}{|c||}{Data Set} & \multicolumn{4}{c||}{Magic Gini Impurity} & \multicolumn{4}{c||}{Information Gain} & \multicolumn{4}{c||}{Weighted Gini Index} & \multicolumn{4}{c|}{Chi-Square} \\ \hline
Name & Rows & Cols & Acc. & Nodes & Depth & T(ms) & Acc. & Nodes & Depth & T(ms) & Acc. & Nodes & Depth & T(ms) & Acc. & Nodes & Depth & T(ms)  \\ \hline
acute & 120 & 7  & 1.0 & \textcolor{blue}{\textbf{4.0}} & 3.3 & 3  & 1.0 & \textcolor{blue}{\textbf{4.0}} & 3.3 &  \textcolor{red}{\textbf{2}}  & 1.0 & 4.2 & 3.3 & \textcolor{red}{\textbf{2}}  & 1.0 & 4.2 & 3.3 & \textcolor{red}{\textbf{2}} \\ \hline
wine & 178 & 14  & \textcolor{magenta}{\textbf{0.97}} & \textcolor{blue}{\textbf{5.8}} & \textcolor{orange}{\textbf{3.9}} & 51  & 0.93 & 8.0 & 4.2 & 59  & 0.89 & 9.8 & 4.5 & \textcolor{red}{\textbf{48}}  & 0.89 & 9.4 & 4.5 & 63 \\ \hline
heart & 270 & 14  & 0.73 & \textcolor{blue}{\textbf{38.4}} & 7.7 & 65  & \textcolor{magenta}{\textbf{0.75}} & 38.8 & 7.0 & \textcolor{red}{\textbf{38}}  & 0.70 & 42.2 & 7.1 & 45  & 0.74 & 40.3 & \textcolor{orange}{\textbf{6.9}} & 63  \\ \hline
ionosphere & 351 & 35  & 0.88 & 19.1 & 7.5 & \textcolor{red}{\textbf{302}}  & 0.88 & \textcolor{blue}{\textbf{19.0}} & 7.0 & 377  & \textcolor{magenta}{\textbf{0.91}} & 22.7 & 7.1 & 413  & 0.89 & 21.5 & \textcolor{orange}{\textbf{6.4}} & 512 \\ \hline
kidney & 400 & 25  & \textcolor{magenta}{\textbf{1.0}} & \textcolor{blue}{\textbf{7.1}} & 4.6 & \textcolor{red}{\textbf{34}}  & 0.98 & 8.0 & \textcolor{orange}{\textbf{4.3}} & 53  & 0.97 & 11.6 & 5.1 & 61  & 0.98 & 10.3 & 4.8 & 87 \\ \hline
voting & 435 & 17  & 0.94 & 24.2 & 6.9 & 28  & \textcolor{magenta}{\textbf{0.95}} & \textcolor{blue}{\textbf{23.9}} & 6.4 & 27  & 0.94 & 25.4 & \textcolor{orange}{\textbf{6.3}} & \textcolor{red}{\textbf{25}}  & 0.94 & 24.9 & \textcolor{orange}{\textbf{6.3}} & 28 \\ \hline
credit-a & 690 & 16  & \textcolor{magenta}{\textbf{0.81}} & 78.2 & 10.4 & 172  & 0.80 & \textcolor{blue}{\textbf{75.4}} & 9.4 & 137  & \textcolor{magenta}{\textbf{0.81}} & 81.9 & 8.8 & \textcolor{red}{\textbf{135}}  & 0.80 & 79.9 & \textcolor{orange}{\textbf{8.5}} & 201 \\ \hline
breast-w & 699 & 10  & 0.93 & 35.9 & 10.1 & \textcolor{red}{\textbf{34}}  & 0.92 & 35.3 & 8.6 & 48  & \textcolor{magenta}{\textbf{0.94}} & \textcolor{blue}{\textbf{34.5}} & 7.9 & 37  & 0.93 & 35.7 & \textcolor{orange}{\textbf{7.8}} & 46 \\ \hline
autism & 704 & 18  & \textcolor{magenta}{\textbf{0.93}} & 46.3 & 7.6 & 53  & 0.92 & \textcolor{blue}{\textbf{46.1}} & \textcolor{orange}{\textbf{7.4}} & \textcolor{red}{\textbf{48}}  & 0.90 & 51.8 & 7.5 & 52  & 0.91 & 48.0 & \textcolor{orange}{\textbf{7.4}} & 57 \\ \hline
parkinson & 765 & 754  & \textcolor{magenta}{\textbf{0.83}} & \textcolor{blue}{\textbf{33.7}} & 10.4 & 22,224  & 0.81 & 34.9 & 7.3 & \textcolor{red}{\textbf{18,253}}  & 0.78 & 42.6 & 8.2 & 30,081  & 0.81 & 40.3 & \textcolor{orange}{\textbf{7.0}} & 37,881 \\ \hline
diabetes & 768 & 9  & 0.69 & 119.5 & 11.1 & 186  & 0.68 & \textcolor{blue}{\textbf{117.4}} & 10.1 & 166  & 0.69 & 125.3 & 9.9 & \textcolor{red}{\textbf{165}}  & \textcolor{magenta}{\textbf{0.70}} & 118.5 & \textcolor{orange}{\textbf{9.2}} & 236 \\ \hline
cars & 1728 & 7  & \textcolor{magenta}{\textbf{1.0}} & \textcolor{blue}{\textbf{47.2}} & 9.9 & \textcolor{red}{\textbf{51}}  & \textcolor{magenta}{\textbf{1.0}} & 47.9 & \textcolor{orange}{\textbf{9.6}} & 53  & 0.99 & 57.1 & \textcolor{orange}{\textbf{9.6}} & 150  & 0.99 & 56.0 & \textcolor{orange}{\textbf{9.6}} & 62 \\ \hline
kr vs. kp & 3196 & 37  & 1.0 & \textcolor{blue}{\textbf{43.3}} & 10.1 & 456  & 1.0 & 43.9 & 10.1 & \textcolor{red}{\textbf{422}}  & 1.0 & 50.8 & \textcolor{orange}{\textbf{9.9}} & 456  & 1.0 & 49.3 & \textcolor{orange}{\textbf{9.9}} & 585 \\ \hline
mushroom & 8124 & 23  & 1.0 & \textcolor{blue}{\textbf{11.0}} & \textcolor{orange}{\textbf{5.1}} & \textcolor{red}{\textbf{972}}  & 1.0 & 11.9 & \textcolor{orange}{\textbf{5.1}} & 1,463  & 1.0 & 14.3 & 5.5 & 1,262  & 1.0 & 12.9 & 5.4 & 1,160 \\ \hline
churn-model & 10000 & 11  & \textcolor{magenta}{\textbf{0.80}} & 1260.9 & 17.2 & \textcolor{red}{\textbf{4,975}}  & \textcolor{magenta}{\textbf{0.80}} & \textcolor{blue}{\textbf{1225.6}} & 15.5 & 5,610  & 0.79 & 1286.2 & 15.1 & 5,432  & 0.79 & 1266.9 & \textcolor{orange}{\textbf{14.2}} & 6,492 \\ \hline
intention & 12330 & 18  & \textcolor{magenta}{\textbf{0.87}} & 884.3 & 19.3 & 7,007  & 0.86 & \textcolor{blue}{\textbf{843.2}} & 15.3 & \textcolor{red}{\textbf{6,886}}  & 0.86 & 921.3 & 14.3 & 8,378  & 0.86 & 884.3 & \textcolor{orange}{\textbf{13.4}} & 8,440 \\ \hline
eeg & 14980 & 15  & 0.84 & \textcolor{blue}{\textbf{1135.7}} & 17.6 & \textcolor{red}{\textbf{11,634}}  & \textcolor{magenta}{\textbf{0.85}} & 1141.9 & 15.9 & 11,820  & 0.84 & 1290.5 & 15.9 & 12,266  & 0.84 & 1236.5 & \textcolor{orange}{\textbf{13.8}} & 11,757 \\ \hline
credit card & 30000 & 24  & 0.73 & 4011.9 & 27.3 & 68,232  & 0.73 & \textcolor{blue}{\textbf{3895.5}} & 22.5 & 62,112  & 0.73 & 4241.7 & 21.8 & \textcolor{red}{\textbf{58,831}}  & 0.73 & 4101.6 & \textcolor{orange}{\textbf{18.7}} & 78,661 \\ \hline
adult & 32561 & 15  & 0.82 & 4273.5 & 29.2 & \textcolor{red}{\textbf{38,266}}  & 0.82 & \textcolor{blue}{\textbf{4064.6}} & 24.7 & 40,943  & 0.82 & 4220.0 & 22.8 & 47,001  & 0.82 & 4173.7 & \textcolor{orange}{\textbf{20.5}} & 46,194 \\ \hline\hline
average & 6226 & 56  & \textcolor{magenta}{\textbf{0.88}} & 635.8 & 11.5 & 8,144  & \textcolor{magenta}{\textbf{0.88}} & \textcolor{blue}{\textbf{615.0}} & 10.2 & \textcolor{red}{\textbf{7,817}}  & 0.87 & 659.7 & 10.0 & 8,676  & 0.87 & 642.9 & \textcolor{orange}{\textbf{9.3}} & 10,133 \\ \hline
\end{tabular}
\caption{Decision tree with different heuristics on various Datasets}
\label{tbl:trees}
\end{table*}

\section{The FOLD-SE Algorithm}

The FOLD-SE algorithm is a top-down rule-learning algorithm. It starts with the candidate rule 

{\tt p(...) :- true.}

\noindent where {\tt p(...)} is the target predicate to learn. It then extends the body with a selected literal (predicate) from among the features so as to cover maximum number of positive examples and avoid covering maximum number of negative examples. The process of selecting a literal to add to the body of the rule relies on heuristics. Traditionally, the Information Gain (IG) heuristics has been used \cite{foil,fold}. It was pioneered in the FOIL algorithm \cite{foil} and adapted by Shakerin et al for their FOLD algorithm \cite{fold}. It is also used in our FOLD-R++ algorithm. We have developed another heuristics based on Gini Impurity \cite{gini-impurity} that FOLD-SE uses. This new heuristics leads to significant reduction in the number of learned default rules and literals, as our experiments, reported later, show. FOLD-SE learns a default theory, so in the next step of learning, it swaps the remaining uncovered positive and negative examples, and recursively applies the literal selection step to learn the exception to the default. Literal selection with swapping of examples continues until reasonable accuracy is obtained. In FOLD-SE, the literal selection process is refined to avoid the \textit{long-tail effect}, explained later. Thus, given the following information, represented as predicates:

\begin{verbatim}
     bird(tweety).         bird(woody).
     cat(kitty).           penguin(polly).
     bird(polly).
\end{verbatim} 

\noindent and given positive examples: 

\begin{verbatim}
    E+: flies(tweety).          flies(woody).
\end{verbatim}

\noindent and negative examples

\begin{verbatim}
    E-: flies(polly).           flies(kitty).
\end{verbatim}

\noindent FOLD, FOLD-R++, and FOLD-SE algorithms will learn the default theory:

{\tt flies(X) :- bird(X), not abnormal(X).

abnormal(X) :- penguin(X).}

\smallskip
\noindent Note that the information above is shown in predicate form for ease of explaining. In reality, the information will have to be represented as a single table with {\tt bird}, {\tt cat}, and {\tt penguin} as features and {\tt flies} as a label. The critical insight in the design of FOLD-SE algorithm is the heuristics used for literal selection. FOLD-SE uses heuristics that is based on computing Gini Impurity, as opposed to Information Gain, which leads to scalable explainability. Our Gini Impurity-based heuristics is discussed next. 

\subsection{Heuristics for Literal Selection}

Most top-down RBML algorithms employ heuristics to guide the literal selection process; information gain (IG) and its variations are the most popular ones. Every selected literal leads to a split on input examples during training. The heuristic used in split-based classifiers greatly impacts the accuracy and structure of the learned model, whether its rule-based or decision tree-based. Specifically, the heuristic used in literal selection of RBML algorithms impacts the number of generated rules and literals, therefore it has an impact on explainability.  

FOLD-SE employs a heuristic that we have devised called Magic Gini Impurity (MGI). MGI is inspired by the Gini Impurity \cite{gini-impurity} heuristics to guide the literal selection process. It helps reduce the number of generated rules and literals while maintaining competitive performance compared to using information gain (IG). 

Gini Impurity for binary classification is defined as:

\begin{equation}
\footnotesize
    GI(tp,fn,tn,fp)=\frac{tp\times fp}{(tp+fp)^2}+\frac{tn\times fn}{(tn+fn)^2}
\label{eq:mgi}
\end{equation}

\noindent where tp, tn, fp, and fn stand for number of true positives, true negatives, false positives, and false negatives, respectively.

%

\noindent
The interesting thing to note is that the Gini Impurity heuristics does not lead to any improvement for the decision tree machine learning algorithm. However, for the FOLD family of algorithms, that learn default rules, the improvement is quite dramatic. We demonstrate this through experiments, next.

To study MGI's effectiveness, we performed a comparison experiment of 4 different heuristics (MGI, information gain, weighted Gini Index, Chi-Square) with 2-way split decision trees on various datasets. 
The decision tree employs the literal selection process of our FOLD-R++ algorithm \cite{foldrpp} for splitting nodes. The accuracy, numbers of generated tree nodes, the average depth of leaf nodes, and time consumption of the 10-fold cross-validation test are averaged and reported in Table \ref{tbl:trees}. This comparison experiment is performed on a small form factor desktop with an Intel i7-8705G CPU and 16 GB RAM. As we can see from the results shown in Table \ref{tbl:trees}, 
MGI heuristic's performance is similar to that for other heuristics. The numbers of generated tree nodes and  average depth of leaf nodes are also very close for all the heuristics. Thus, for 2-way split decision trees, the MGI heuristics does not provide any significant benefit.

\begin{table*}[]
\centering
\setlength{\tabcolsep}{2pt}
\begin{tabular}{|l|c|c||c|c|c|c|c|c|c||c|c|c|c|c|c|c|}
\hline
\multicolumn{3}{|c||}{Data Set} & \multicolumn{7}{c||}{FOLD-R++} & \multicolumn{7}{c|}{FOLD-R++ with MGI} \\ \hline
Name & Rows & Cols & Acc & Prec & Rec & F1 & T(ms) & Rules & Preds & Acc & Prec & Rec & F1 & T(ms) & Rules & Preds \\ \hline
acute & 120 & 7 & 0.99 & 1.0 & 0.99 & 0.99 & 2 & 2.7$\pm$0.46  & 3.0$\pm$0.0    & 0.99 & 1.0 & 0.99 & 0.99 & \textcolor{orange}{\textbf{1}} & 2.7$\pm$0.46  & 3.0$\pm$0.0  \\ \hline
heart & 270 & 14 & \textcolor{magenta}{\textbf{0.77}} & \textcolor{teal}{\textbf{0.80}} & \textcolor{cyan}{\textbf{0.80}} & \textcolor{violet}{\textbf{0.79}} & 38 & 15.9$\pm$4.6 & 32.2$\pm$12.0    & 0.74 & 0.77 & 0.78 & 0.77 & \textcolor{orange}{\textbf{11}} & \textcolor{blue}{\textbf{4.3$\pm$3.3}}  & \textcolor{red}{\textbf{8.9$\pm$8.2}}  \\ \hline
ionosphere & 351 & 35 & 0.90 & \textcolor{teal}{\textbf{0.92}} & 0.93 & 0.92 & 275 & 12.4$\pm$0.9 & 19.7$\pm$0.5    & \textcolor{magenta}{\textbf{0.91}} & 0.89 & \textcolor{cyan}{\textbf{0.98}} & \textcolor{violet}{\textbf{0.93}} & \textcolor{orange}{\textbf{94}} & \textcolor{blue}{\textbf{5.1$\pm$0.7}}  & \textcolor{red}{\textbf{7.1$\pm$0.7}}  \\ \hline
kidney & 400 & 25 & \textcolor{magenta}{\textbf{0.99}} & 1.0 & \textcolor{cyan}{\textbf{0.98}} & \textcolor{violet}{\textbf{0.99}} & \textcolor{orange}{\textbf{16}} & \textcolor{blue}{\textbf{4.9$\pm$0.3}} & \textcolor{red}{\textbf{5.9$\pm$0.5}}     & 0.98 & 1.0 & 0.97 & 0.98 & 25 & 5.2$\pm$0.6 & 6.3$\pm$0.6 \\ \hline
voting & 435 & 17 & 0.94 & 0.92 & 0.93 & 0.92 & \textcolor{orange}{\textbf{23}} & 10.0$\pm$2.1 & 27.2$\pm$6.6   & \textcolor{magenta}{\textbf{0.95}} & 0.92 & \textcolor{cyan}{\textbf{0.96}} & \textcolor{violet}{\textbf{0.94}} & 25 & \textcolor{blue}{\textbf{7.8$\pm$2.2}} & \textcolor{red}{\textbf{20.2$\pm$6.7}} \\ \hline
credit-a & 690 & 16 & 0.83 & 0.90 & 0.78 & 0.83 & 84 & 10.3$\pm$3.8 & 23.3$\pm$9.8   & \textcolor{magenta}{\textbf{0.85}} & \textcolor{teal}{\textbf{0.93}} & 0.78 & \textcolor{violet}{\textbf{0.85}} & \textcolor{orange}{\textbf{39}} & \textcolor{blue}{\textbf{2.2$\pm$1.2}} & \textcolor{red}{\textbf{3.8$\pm$2.9}} \\ \hline
breast-w & 699 & 10 & 0.95 & 0.97 & \textcolor{cyan}{\textbf{0.95}} & 0.96 & 34 & 10.5$\pm$1.9 & 18.6$\pm$5.4  & 0.95 & \textcolor{teal}{\textbf{0.98}} & 0.94 & 0.96 & \textcolor{orange}{\textbf{32}} & \textcolor{blue}{\textbf{8.1$\pm$2.0}} & \textcolor{red}{\textbf{12.9$\pm$3.4}} \\ \hline
autism & 704 & 18 & \textcolor{magenta}{\textbf{0.93}} & 0.95 & \textcolor{cyan}{\textbf{0.95}} & \textcolor{violet}{\textbf{0.95}} & 62 & 25.4$\pm$1.8 & 54.8$\pm$5.7   & 0.92 & 0.95 & 0.94 & 0.94 & \textcolor{orange}{\textbf{41}} & \textcolor{blue}{\textbf{19.4$\pm$4.4}} & \textcolor{red}{\textbf{43.4$\pm$10.8}} \\ \hline
parkinson & 765 & 754 & \textcolor{magenta}{\textbf{0.82}} & \textcolor{teal}{\textbf{0.85}} & 0.93 & 0.89 & 10,757 & 13.7$\pm$4.2 & 21.2$\pm$7.5  & 0.81 & 0.82 & \textcolor{cyan}{\textbf{0.96}} & 0.89 & \textcolor{orange}{\textbf{7,469}} & \textcolor{blue}{\textbf{7.4$\pm$1.6}} & \textcolor{red}{\textbf{13.2$\pm$4.2}} \\ \hline
diabetes & 768 & 9 & 0.74 & \textcolor{teal}{\textbf{0.79}} & 0.83 & 0.80 & 66 & 8.3$\pm$3.2 & 19.4$\pm$8.8  & 0.74 & 0.78 & \textcolor{cyan}{\textbf{0.84}} & \textcolor{violet}{\textbf{0.81}} & \textcolor{orange}{\textbf{48}} & \textcolor{blue}{\textbf{4.8$\pm$2.0}} & \textcolor{red}{\textbf{11.6$\pm$4.6}}  \\ \hline
cars & 1728 & 7 & 0.96 & 1.0 & 0.95 & 0.97 & \textcolor{orange}{\textbf{31}} & 12.3$\pm$2.6 & 29.8$\pm$7.0  & 0.96 & 1.0 & 0.95 & 0.97 & 35 & \textcolor{blue}{\textbf{8.8$\pm$0.8}} & \textcolor{red}{\textbf{17.5$\pm$4.0}} \\ \hline
kr vs. kp & 3196 & 37 & \textcolor{magenta}{\textbf{0.99}} & \textcolor{teal}{\textbf{1.0}} & \textcolor{cyan}{\textbf{0.99}} & \textcolor{violet}{\textbf{0.99}} & 226 & 19.3$\pm$3.8 & 46.7$\pm$11.0   & 0.97 & 0.97 & 0.97 & 0.97 & \textcolor{orange}{\textbf{170}} & \textcolor{blue}{\textbf{11.4$\pm$2.0}} & \textcolor{red}{\textbf{27.6$\pm$7.7}} \\ \hline
mushroom & 8124 & 23 & 1.0 & 1.0 & 1.0 & 1.0 & 281 & 7.9$\pm$0.3 & \textcolor{red}{\textbf{11.9$\pm$0.3}}   & 1.0 & 1.0 & 1.0 & 1.0 & \textcolor{orange}{\textbf{257}} & \textcolor{blue}{\textbf{7.0$\pm$0.0}} & 13.4$\pm$0.8 \\ \hline
intention & 12330 & 18 & 0.90 & 0.95 & 0.93 & 0.94 & 1,085 & 8.4$\pm$5.5 & 23.0$\pm$15.8  & 0.90 & 0.95 & 0.93 & 0.94 & \textcolor{orange}{\textbf{691}} & \textcolor{blue}{\textbf{4.4$\pm$1.7}} & \textcolor{red}{\textbf{11.4$\pm$5.2}} \\ \hline
eeg & 14980 & 15 & \textcolor{magenta}{\textbf{0.72}} & \textcolor{teal}{\textbf{0.76}} & \textcolor{cyan}{\textbf{0.72}} & \textcolor{violet}{\textbf{0.74}} & 2,735 & 69.1$\pm$17.3 & 152.6$\pm$35.6  & 0.68 & 0.75 & 0.64 & 0.69 & \textcolor{orange}{\textbf{1,353}} & \textcolor{blue}{\textbf{18.7$\pm$7.9}} & \textcolor{red}{\textbf{40.8$\pm$19.5}} \\ \hline
credit card & 30000 & 24 & 0.82 & 0.83 & 0.96 & 0.89 & 5,954 & 19.1$\pm$6.4 & 48.8$\pm$14.7  & 0.82 & 0.83 & 0.96 & 0.89 & \textcolor{orange}{\textbf{3,827}} & \textcolor{blue}{\textbf{10.9$\pm$3.2}} & \textcolor{red}{\textbf{25.5$\pm$8.7}} \\ \hline
adult & 32561 & 15 & 0.84 & 0.86 & 0.95 & 0.90 & 2,508 & 16.8$\pm$2.4 & 46.7$\pm$10.4   & 0.84 & 0.86 & 0.95 & 0.90 & \textcolor{orange}{\textbf{1,414}} & \textcolor{blue}{\textbf{6.8$\pm$1.5}} & \textcolor{red}{\textbf{13.2$\pm$3.6}} \\ \hline
rain in aus & 145460 & 24 & 0.79 & \textcolor{teal}{\textbf{0.87}} & 0.84 & 0.86 & 26,203 & 48.2$\pm$10.6 & 115.8$\pm$39.1   & \textcolor{magenta}{\textbf{0.80}} & 0.86 & \textcolor{cyan}{\textbf{0.87}} & 0.86 & \textcolor{orange}{\textbf{15,003}} & \textcolor{blue}{\textbf{14.6$\pm$4.7}} & \textcolor{red}{\textbf{57.9$\pm$31.0}} \\ \hline \hline
average & 14087 & 59 & 0.88 & \textcolor{teal}{\textbf{0.91}} & 0.91 & \textcolor{violet}{\textbf{0.91}} & 2,704 & 17.5$\pm$4.0 & 38.9$\pm$10.6   & 0.88 & 0.90 & 0.91 & 0.90 & \textcolor{orange}{\textbf{1,493}} & \textcolor{blue}{\textbf{8.3$\pm$2.2}} & \textcolor{red}{\textbf{18.8$\pm$6.8}} \\ \hline

\end{tabular}
\caption{Comparison of Magic Gini impurity (MGI) and information gain (IG) in FOLD-R++}
\label{tbl:foldrppmgi}
\end{table*}

Next, we use Magic Gini Impurity (MGI) as the heuristics in the FOLD-R++ algorithm and compare it to FOLD-R++ with the Information Gain (IG) heuristics. \textit{Interestingly, the number of rules and predicates in the rules is reduced drastically.} As we can see in Table \ref{tbl:foldrppmgi}, the number of generated rules and generated literals is reduced significantly with MGI compared to IG, while prediction performance remains the same. Note that results reported are for 10-fold cross-validation test, thus, the number of rules can be fractional due to computing the average. Standard deviation is also reported (avg {$\pm$} s.d.). Thus, Magic Gini Impurity significantly improves interpretability and explainability of the FOLD-R++ algorithm, that aims to learn default rules, while preserving performance. As noted earlier, interestingly, use of our MGI heuristics in decision trees (that learn formulas in conjunctive normal form) does not produce much benefit. Why MGI benefits one and not the other is a subject of further study.

\subsection{Comparison of Feature Values}

\begin{table}[]
\centering
\begin{tabular}{|c|c||c|c|}
        \cline{1-4}
        \textbf{comparison}  & \textbf{evaluation} & \textbf{comparison}  & \textbf{evaluation} \\
        \cline{1-4}
        10 $=$ `cat' & False & 10 $\neq$ `cat' & True\\ 
        \cline{1-4}
        10 $\le$ `cat' & False & 10 $>$ `cat' & False \\ 
        \cline{1-4}
        10 $\nleq$ `cat' & True & 10 $\ngtr$ `cat' & True\\ 
        \cline{1-4}
        \end{tabular}
\caption{Comparing numerical and categorical values }
\label{tbl:example1}
\end{table}

During the learning process, FOLD-SE has to compare categorical and numerical data values. FOLD-SE employs a carefully designed comparison strategy, which is an extension of the comparison strategy of our FOLD-R++ algorithm for comparing categorical and numerical values. This gives FOLD-SE the ability to  elegantly handle mixed-type values and, thus, learn from datasets that may have features containing both numerical and categorical values  (a missing value is considered as a categorical value). The comparison between two numerical values or two categorical values in FOLD-R++ is straightforward, as commonsense would dictate, i.e., two numerical (resp. categorical) values are equal if they are identical, otherwise they are unequal. The equality between a numerical value and a categorical value is always false, and the inequality between a numerical value and a categorical value is always true. In addition, numerical comparisons ($\leq$ and $>$) between a numerical value and a categorical value is always false. However, the numerical comparisons $\leq$ and $>$ are not complementary to each other with this comparison assumption. For example, $x$ $\leq$ 4 means that $x$ is a number and $x$ is less than or equal to 4. The opposite of $x$ $\leq$ 4 should be $x$ is a number greater than 4 \textit{or} $x$ is not a number. 
Without the opposite of these two numerical comparisons being used, the literal selection process 
would be limited. The FOLD-SE algorithm, thus, extends the comparison operation with $\nleq$ and $\ngtr$ as the opposites of $\leq$ and $>$, respectively. The literals with $\nleq$ and $\ngtr$ will be candidate literals in the literal selection process but converted to their opposites, $\leq$ and $>$, in the final results. An example is shown in Table \ref{tbl:example1}. 

Given E$^+$={\tt \{3,4,4,5,x,x,y\}}, E$^-$={\tt \{1,1,1,2,3,y,y,z\}}, and 
literal ${\tt (i,>,3)}$ in Table \ref{tbl:example2}, the true positive example E$_{\tt tp}$, false negative examples E$_{\tt fn}$, true negative examples E$_{\tt tn}$, and false positive examples E$_{\tt fp}$ implied by the literal are {\tt \{4,4,5\}}, {\tt \{3,x,x,y\}}, {\tt \{1,1,1,2,3,y,y,z\}}, {\tt \O} respectively. Then, the heuristic of literal ${\tt (i,>,3)}$ is calculated as MGI$_{\tt (i,>,3)}${\tt (3,4,8,0)}=-0.38 through Formula (\ref{eq:mgi}) above.

\begin{table}[]
\centering
\setlength{\tabcolsep}{1pt}
\begin{tabular}{|c|c|c|}
        \cline{1-3}
          & \multicolumn{1}{c|}{\textbf{i$^{th}$ feature values}} & \textbf{count} \\
        \cline{1-3}
        $\mathbf{E^+}$ & 3 4 4 5 x x y & \textbf{7} \\
        \cline{1-3}
        $\mathbf{E^-}$ & 1 1 1 2 3 y y z & \textbf{8} \\
        \cline{1-3}
        \hline \hline
        $\mathbf{E_{tp(i,>,3)}}$ & 4 4 5 & \textbf{3} \\
        \cline{1-3}
        $\mathbf{E_{fn(i,>,3)}}$ & 3 x x y & \textbf{4} \\
        \cline{1-3}
        $\mathbf{E_{tn(i,>,3)}}$ & 1 1 1 2 3 y y z & \textbf{8} \\
        \cline{1-3}
        $\mathbf{E_{fp(i,>,3)}}$ & \O & \textbf{0} \\
        \cline{1-3}
        \end{tabular}
\caption{Evaluation and count for literal$(i,>,3)$. }
\label{tbl:example2}
\end{table}

\subsection{Literal Selection}

The FOLD-R++ algorithm starts the learning process with the candidate rule {\tt p(\ldots):- true.}, where {\tt p(\ldots)} is the target predicate to learn. It specializes the rule by adding literals to its body during the training process. It adds a literal that maximizes information gain. FOLD-SE extends the literal selection process of FOLD-R++ by employing MGI as a heuristic instead of IG. In addition, candidate literals of the form ${\tt m \nleq n}$ and ${\tt m \ngtr n}$ are also considered. 
The literal selection process of FOLD-SE is summarized in Algorithm \ref{algo:selection}. In line 2, ${\tt cnt^+}$ and ${\tt cnt^-}$ are dictionaries that hold, respectively, the numbers of positive and negative examples of each unique value. In line 3, ${\tt set_n}$, ${\tt set_c}$ are sets that hold, respectively, the unique numerical and categorical values. In line 4, ${\tt tot_{n}^{+}}$ and ${\tt tot_{n}^{-}}$ are the total number of, respectively, positive and negative examples with numerical values; ${\tt tot_{c}^{+}}$ and ${\tt tot_{c}^{-}}$ are the total number of, respectively, positive and negative examples with categorical values. In line 6, the prefix sums of numerical values have been computed as preparation for calculating heuristics of candidate literals. After the prefix sum calculation process, ${\tt cnt^+[x]}$ and ${\tt cnt^-[x]}$ represents the number of positive examples and negative examples that have a value less than or equal to ${\tt x}$. Preparing parameters correctly is essential to calculating MGI values for candidate literals. In line 11, the MGI value for literal ${\tt (i, \le, x)}$ is computed by taking parameters ${\tt cnt^+[x]}$ as number of true positive examples,  ${\tt tot_{n}^{+}-cnt^+[x]+tot_{c}^{+}}$ as the number of false positive examples, ${\tt tot_{n}^{-}-cnt^-[x]+tot_{c}^{-}}$ as the number of true negative examples, and ${\tt cnt^-[x]}$ as the number of false positive examples. The reason for this is as follows: for the literal ${\tt (i,\le,x)}$, only numerical values that are less than or equal to ${\tt x}$ can be evaluated as positive, otherwise negative. ${\tt tot_{n}^{+}-cnt^+[x]+tot_{c}^{+}}$ represents the number of positive examples that have a value greater than ${\tt x}$ plus the total number of positive examples with categorical values. ${\tt tot_{n}^{-}-cnt^-[x]+tot_{c}^{-}}$ represents the number of negative examples that have a value greater than ${\tt x}$ plus the total number of negative examples with categorical values. ${\tt cnt^-[x]}$ represents the number of negative examples that have a value less than or equal to ${\tt x}$. The heuristic calculation for other candidate literals also follows the same comparison regime mentioned above. Finally, the {\tt best\_literal\_on\_attr} function returns the best heuristic score and the corresponding literal except the literals that have been used in current rule-learning process. 

\begin{algorithm}[!h]
\caption{FOLD-SE Find Best Literal function}
\label{algo:selection}
\SetKwInOut{Input}{input}
\SetKwInOut{Output}{output}

\SetKwFunction{cntclass}{count\_class}
\SetKwFunction{cnttotal}{count\_total}
\SetKwFunction{unique}{unique\_values}
\SetKwFunction{cntsort}{counting\_sort}
\SetKwFunction{h}{MGI}
\SetKwFunction{bestpair}{best\_pair}
\SetKwFunction{bestonattr}{best\_literal\_on\_attr}
\SetKwFunction{bestliteral}{find\_best\_literal}


\DontPrintSemicolon

\Input{$E^+$: positive examples, $E^-$: negative examples, $used$: used literals}
\Output{$literal$: the literal that has best heuristic score}
\SetKwProg{Fn}{Function}{}{end}

\Fn{\bestonattr{$E^+,E^-,i,used$}}{
    $cnt^+,cnt^-\gets$ \cntclass{$E^+,E^-,i$}\;
    $set_n,set_c\gets$ \unique{$E^+,E^-,i$}\;
    $tot_{n}^{+},tot_{n}^{-},tot_{c}^{+},tot_{c}^{-} \gets$ \cnttotal{$E^+,E^-,i$}\;
    $num\gets$ \cntsort{$set_n$}\;
    \For{$i\leftarrow 1$ \KwTo $|num|$}{
        $cnt^+[num_{j}]\gets cnt^+[num_{j}]+cnt^+[num_{j-1}]$\;
        $cnt^-[num_{j}]\gets cnt^-[num_{j}]+cnt^-[num_{j-1}]$\;
    }
    \For{$x \in set_n$}{
        $score[(i,\leq,x)] \gets$ \h{$cnt^+[x], tot_{n}^{+}-cnt^+[x]+tot_{c}^{+}, tot_{n}^{-}-cnt^-[x]+tot_{c}^{-}, cnt^-[x]$}\;
        $score[(i,>,x)] \gets$ \h{$tot_{n}^{+}-cnt^+[x], cnt^+[x]+tot_{c}^{+}, cnt^-[x]+tot_{c}^{-}, tot_{n}^{-}-cnt^-[x]$}\;
        $score[(i,\nleq,x)] \gets$ \h{$tot_{n}^{+}-cnt^+[x]+tot_{c}^{+}, cnt^+[x], cnt^-[x], tot_{n}^{-}-cnt^-[x]+tot_{c}^{-}$}\;
        $score[(i,\ngtr,x)] \gets$ \h{$cnt^+[x]+tot_{c}^{+}, tot_{n}^{+}-cnt^+[x], tot_{n}^{-}-cnt^-[x], cnt^-[x]+tot_{c}^{-}$}\;
    }
    \For{$c \in set_s$}{
        $score[(i,=,c)] \gets$ \h{$cnt^+[c], tot_{c}^{+}-cnt^+[c]+tot_{n}^{+}, tot_{c}^{-}-cnt^-[c]+tot_{n}^{-}, cnt^-[c]$}\;
        $score[(i,\ne,c)] \gets$ \h{$tot_{c}^{+}-cnt^+[c]+tot_{n}^{+}, cnt^+[c], cnt^-[c], tot_{c}^{-}- cnt^-[c]+tot_{n}^{-}$}\;
    }
    $h, literal \gets$ \bestpair{$score,used$}\;
    \textbf{return} $h, literal$
}

\Fn{\bestliteral{$E^+,E^-,used$}}{
    $best\_h, literal \gets -\infty, invalid$\;
    \For{$i\leftarrow 1$ \KwTo N}{
        $h, lit \gets$ \bestonattr{{$E^+$},{$E^-$},{$i$},{$used$}}\;
        \If{$best\_h < h$}{
            $best\_h, literal \gets h, lit$\;
        }
    }
    \textbf{return} $literal$
}

\end{algorithm}

\begin{exmp}
\label{ex:pinguin2}
Given positive and negative examples in Table \ref{tbl:example1}, E$^+$, E$^-$, with mixed type of values on i$^{th}$ feature, the target is to find the literal with the best MGI heuristic
on the given feature. There are 7 positive examples, their values on i$^{th}$ feature are $[3,4,4,5,x,x,y]$, and the values on i$^{th}$ feature of the 8 negative examples are $[1,1,1,2,3,y,y,z]$. 

\end{exmp}

\noindent 
With the given examples and specified feature, the number of positive examples and negative examples for each unique value are counted first, which are shown as $\mathbf{count^+}$, $\mathbf{count^-}$ in Table \ref{tbl:example3}. Then, the prefix sum arrays are calculated for computing heuristic as $\mathbf{sum^{+}_{pfx}}$, $\mathbf{sum^{-}_{pfx}}$. Table \ref{tbl:example4} shows the MGI heuristic
for each candidate literal and the literal${\tt (i,\nleq,2)}$ gets selected as it has the highest score.

\begin{table}[]
\centering
\begin{tabular}{|c|c|c|c|c|c|c|c|c|}
        \cline{1-9}
          & \multicolumn{8}{c|}{\textbf{i$^{th}$ feature values}}\\
        \cline{1-9}
        $\mathbf{E^+}$ & 3 & 4 & 4 & 5 & x & x & y & \\
        \cline{1-9}
        $\mathbf{E^-}$ & 1 & 1 & 1 & 2 & 3 & y & y & z  \\
        \cline{1-9}
        \hline \hline
        \textbf{value} & 1 & 2 & 3 & 4 & 5 & x & y & z \\
        \cline{1-9}
        $\mathbf{count^+}$ & 0 & 0 & 1 & 2 & 1 & 2 & 1 & 0 \\
        \cline{1-9}
        $\mathbf{sum^{+}_{pfx}}$ & 0 & 0 & 1 & 3 & 4 & NA & NA & NA \\
        \cline{1-9}
        $\mathbf{count^-}$ & 3 & 1 & 1 & 0 & 0 & 0 & 2 & 1 \\
        \cline{1-9}
        $\mathbf{sum^{-}_{pfx}}$ & 3 & 4 & 5 & 5 & 5 & NA & NA & NA \\
        \cline{1-9}
        \end{tabular}
\caption{Top: Examples and values on i$^{th}$ feature. Bottom: positive/negative count and prefix sum on each value }
\label{tbl:example3}
\end{table}

\begin{table}[]
\centering
{
\setlength{\tabcolsep}{1pt}
\begin{tabular}{|c|c|c|c|c|c|c|c|c|}
\cline{1-9}
 & \multicolumn{8}{c|}{\textbf{heuristic}}  \\
\cline{1-9}
\textbf{value} & 1 & 2 & 3 & 4 & 5 & x & y & z \\
\cline{1-9}
\textbf{$\leq$ value} & $-\infty$ & $-\infty$ & $-\infty$ & $-\infty$ & $-\infty$ & NA & NA & NA \\
\cline{1-9}
\textbf{$>$ value} & -0.47 & -0.44 & -0.38 & -0.46 & -0.50 & NA & NA & NA \\
\cline{1-9}
\textbf{$\nleq$ value} & -0.39 & \textbf{-0.35} & -0.43 & -0.49 & -0.50 & NA & NA & NA \\
\cline{1-9}
\textbf{$\ngtr$ value} & $-\infty$  & $-\infty$ & $-\infty$ & $-\infty$ & $-\infty$ & NA & NA & NA \\
\cline{1-9}
\textbf{$=$ value} & NA & NA & NA & NA & NA & -0.42 & $-\infty$ & $-\infty$  \\
\cline{1-9}
\textbf{$\ne$ value} & NA & NA & NA & NA & NA & $-\infty$ & -0.49 & -0.47 \\
\cline{1-9}
\end{tabular}}
\caption{The heuristic on i$^{th}$ feature with given examples}
\label{tbl:example4}
\end{table}

\subsection{Rule Pruning}

Our FOLD-R++ algorithm \cite{foldrpp} is a recent rule-based machine learning algorithm for binary classification that generates a normal logic program in which all the default rules have the same rule head (target predicate). An example is covered means that it is predicted as positive. An example covered by any default rule in the set would imply the rule head is true. The FOLD-R++ algorithm generates a model by learning one rule at a time. After learning a rule, the already covered examples would be ruled out for better literal selection of remaining examples. If the $ratio$ of false positive examples to true positive examples drops below the preset threshold, it would next learn exceptions by swapping remaining positive and negative examples then calling itself recursively. The $ratio$ stands for the upper bound on the number of true positive examples to the number of false positive examples implied by the default part of a rule. It helps speed up the training process and reduces the number of rules learned. The training process of FOLD-R++ is also a process of ruling out already covered examples. Later generated rules cover fewer examples than the early generated ones. In other words, FOLD-R++  suffers from \textit{long-tail effect}. Here is an example:

\begin{exmp}
\label{ex:adult1}
The ``Adult Census Income" is a classical classification task that contains 32,561 records. We treat 80\% of the data as training examples and 20\% as testing examples. The task is to learn the income status of individuals (more/less than 50K/year) based on features such as gender, age, education, marital status, etc. The FOLD-R++ algorithm generates the following program that contains 9 rules: 
\end{exmp}

{
\small
\begin{verbatim}
(1) income(X,'<=50K') :- 
 [3428] not marital_status(X,'Married-civ-spouse'), 
        not ab3(X,'True'). 
(2) income(X,'<=50K') :- 
 [1999] marital_status(X,'Married-civ-spouse'), 
        education_num(X,N1), N1=<12.0, capital_gain(X,N2), 
        N2=<5013.0, not ab5(X,'True'), not ab6(X,'True').
(3) income(X,'<=50K') :-  occupation(X,'Farming-fishing'), 
 [1]    workclass(X,'Self-emp-not-inc'), 
        education_num(X,N1), N1>12.0, capital_gain(X,N2), 
        N2>5013.0.  
(4) ab1(X,'True') :- not workclass(X,'Local-gov'), 
 [2]    capital_gain(X,N2), N2=<7978.0, education_num(X,N1), 
        N1=<10.0.
(5) ab2(X,'True') :- capital_gain(X,N2), N2>27828.0, 
 [0]    N2=<34095.0.
(6) ab3(X,'True') :- capital_gain(X,N2), N2>6849.0, 
 [0]    age(X,N3), N3>20.0, not ab1(X,'True'), 
        not ab2(X,'True').
(7) ab4(X,'True') :- workclass(X,'Local-gov'), 
 [0]    native_country(X,'United-States').
(8) ab5(X,'True') :- not race(X,'Amer-Indian-Eskimo'), 
 [0]    education_num(X,N1), N1=<8.0, capital_loss(X,N4), 
        N4>1735.0, N4=<1902.0, not ab4(X,'True').
(9) ab6(X,'True') :- occupation(X,'Tech-support'), 
 [0]    not education(X,'11th'), education_num(X,N1), 
        N1>5.0, N1=<8.0, age(X,N3), N3=<36.0.
\end{verbatim}  
}

\noindent
The above generated rules achieve 0.85 accuracy and 0.90 F$_1$ score. The first rule covers 3428 test examples and the second rule covers 1999 test examples. 
Subsequent rules only cover small number of test examples. This long-tail effect is due to the overfitting on the training data. Our FOLD-SE algorithm introduces a hyper-parameter \textit{tail} to limit the minimum number/percentage of training examples that a rule can cover. It helps reduce the number of generated rules and generated literals by reducing overfitting of outliers. This rule pruning is not a post-process after training, rather rules are pruned during the training process itself which helps speed-up training. With the $tail$ parameter, FOLD-SE can be easily tuned to obtain a trade-off between accuracy and explainability. The FOLD-SE algorithm is summarized in Algorithm \ref{algo:foldse}. The added rule pruning process is carried out in Line 33--38 of Algorithm \ref{algo:foldse}. When a learned rule cannot cover enough training examples, the {\tt learn\_rule} function returns. Except for the input parameter \textit{tail} and the rule pruning process, FOLD-R++ and FOLD-SE have the same algorithmic framework. Of course, the heuristic used in FOLD-SE is MGI, while FOLD-R++ uses IG.

\begin{algorithm}[!h]
\caption{FOLD-SE Algorithm}
\label{algo:foldse}
\SetKwInOut{Input}{input}
\SetKwInOut{Output}{output}
\SetKwFunction{find}{find\_best\_literl}
\SetKwFunction{foldse}{learn\_rule\_set}
\SetKwFunction{learn}{learn\_rule}
\SetKwFunction{cover}{cover}
\DontPrintSemicolon

\Input{$E^+$: positive examples, $E^-$: negative examples, $used$: used literals, $ratio$: exception ratio, $tail$: covering limit}
\Output{$R=\{r_1,...,r_n\}$: a set of default rules with exceptions}
\SetKwProg{Fn}{Function}{}{end}

\Fn{\foldse{$E^+,E^-,used,tail$}}{
    $R \gets $ \O\;
    \While{$|E^+| > 0$}{
        $r \gets$ \learn{${E^+},{E^-},{used},{tail}$}\;
        $E_{\textit{fn}} \gets$ \cover{$r,E^+,\textit{false}$}\;
        \If{$|E_{\textit{fn}}|=|E^+|$}{\textbf{break}}
        $E^+ \gets E_{\textit{fn}}$\;
        $R \gets R \cup r$\;
    }
    \textbf{return} $R$
}
\Fn{\learn{${E^+},{E^-},{used},{tail}$}}{
    $L\gets $ \O\;
    $r\gets $ \textbf{$rule_{\empty}$}\;
    \While{\textbf{true}}{
        $l\gets$ \find{${E^+},{E^-},{used}$}\;
        $L\gets L \cup l$\;
        $r.\textit{default} \gets L$\;
        $E^+ \gets$ \cover{$r,\ E^+,\ true$}\;
        $E^- \gets$ \cover{$r,\ E^-,\ true$}\;
        \If{$l$ is invalid \textbf{or} $|E^-| \leq |E^+| * ratio$}{
            \eIf{$l$ is invalid}{
                $r.\textit{default} \gets L$
            }
            {
                $ab \gets$ \foldse{${E^-},{E^+},{used+L},{tail}$}\;
                $r.exception \gets ab$
            }
            \textbf{break}
        }
    }
    \If{$tail>0$}{
        $E^+ \gets$ \cover{$r,\ E^+,\ true$}\;
        \If{$|E^+|<tail$}{
            \textbf{return} invalid
        }
    }
    \textbf{return} $r$
}
\end{algorithm}

\subsection{Complexity Analysis}

Next, we analyze the complexity of our FOLD-SE algorithm.
If $M$ is the number of training examples and N is the number of features that have been included in the training, the time complexity of finding the best literal of a feature is $O(M)$, assuming that counting sort is used at line 5 in Algorithm \ref{algo:selection}. Therefore, the complexity of finding the best literal of all features is $O(MN)$. The worst training case is that each generated rule only covers one training example and each literal only help exclude one example. In this case, $O(M^2)$ literals would be selected in total. Hence, the worst case time complexity of FOLD-SE is $O(M^3N)$. 

Additionally, it is straightforward to prove that the FOLD-SE algorithm always terminates. 
The {\tt learn\_rule\_set} function calls the {\tt learn\_rule} function to generate rules that cover target examples till all the target examples have been covered or the learned rule is invalid. The while-loop in the {\tt learn\_rule\_set} iterate at most $|E^+|$ times while excluding the already covered examples. The {\tt learn\_rule} function specializes the rule by adding the best literal to the rule body. By adding literals to the rules, the numbers of true positive and false positive examples the rule implies can only monotonically decrease. The learned valid literal excludes at least one false positive example that the rule implies. So the while-loop in the {\tt learn\_rule} function iterates at most $|E^-|$ times. When the $|E^-|<|E^+|*ratio$ condition is triggered, the {\tt learn\_rule\_set} function is called to learn exception rules for the current default rule. Finally, the number of iteration of the for-loop in {\tt find\_best\_literal} function is finite. Consequently, The FOLD-SE algorithm always terminates.

\begin{table*}[!h]
\centering
\setlength{\tabcolsep}{2pt}
\small
\begin{tabular}{|l|c|c||c|c|c|c|c||c|c|c|c|c||c|c|c|c|c|}
\hline
\multicolumn{3}{|c||}{Data Set} & \multicolumn{5}{c||}{RIPPER} & \multicolumn{5}{c||}{FOLD-R++} & \multicolumn{5}{c|}{FOLD-SE} \\ \hline
Name & Rows & Cols & Acc & F1 & T(ms) &Rules &Preds & Acc & F1 & T(ms) &Rules &Preds & Acc & F1 & T(ms) &Rules &Preds \\ \hline
acute & 120 & 7 & 0.93 & 0.92 & 95 & \textcolor{blue}{\textbf{2.0$\pm$0.0}} & 4.0$\pm$0.0 & 0.99 & 0.99 & 2 & 2.7$\pm$0.5 & \textcolor{red}{\textbf{3.0$\pm$0.0}} & \textcolor{magenta}{\textbf{1.0}} & \textcolor{violet}{\textbf{1.0}} & \textcolor{orange}{\textbf{1}} & \textcolor{blue}{\textbf{2.0$\pm$0.0}} & \textcolor{red}{\textbf{3.0$\pm$0.0}}\\ \hline
heart & 270 & 14 & 0.76 & 0.77 & 317 & 5.4$\pm$0.9 & 12.9$\pm$2.8 & \textcolor{magenta}{\textbf{0.77}} & \textcolor{violet}{\textbf{0.79}} & 38 & 15.9$\pm$4.6 & 32.2$\pm$12.0 & 0.74 & 0.77 & \textcolor{orange}{\textbf{13}} & \textcolor{blue}{\textbf{4.0$\pm$3.0}} & \textcolor{red}{\textbf{9.1$\pm$8.5}} \\ \hline
ionosphere & 351 & 35 & 0.72 & 0.73 & 1,161 & 8.5$\pm$4.1 & 13.9$\pm$6.0 & 0.90 & 0.92 & 275 & 12.4$\pm$0.9 & 19.7$\pm$0.5 & \textcolor{magenta}{\textbf{0.91}} & \textcolor{violet}{\textbf{0.93}} & \textcolor{orange}{\textbf{119}} & \textcolor{blue}{\textbf{3.6$\pm$0.5}} & \textcolor{red}{\textbf{7.1$\pm$0.7}} \\ \hline
kidney & 400 & 25 & 0.98 & 0.98 & 750 & 7.1$\pm$0.5 & 8.5$\pm$0.8 & 0.99 & 0.99 & \textcolor{orange}{\textbf{16}} & \textcolor{blue}{\textbf{4.9$\pm$0.3}} & \textcolor{red}{\textbf{5.9$\pm$0.5}} & \textcolor{magenta}{\textbf{1.0}} & \textcolor{violet}{\textbf{1.0}} & \textcolor{orange}{\textbf{16}} & \textcolor{blue}{\textbf{4.9$\pm$0.3}} & 6.1$\pm$0.3 \\ \hline
voting & 435 & 17 & \textcolor{magenta}{\textbf{0.95}} & 0.92 & 172 & \textcolor{blue}{\textbf{4.1$\pm$1.4}} & \textcolor{red}{\textbf{8.9$\pm$4.1}} & 0.94 & 0.92 & 23 & 10.0$\pm$2.1 & 27.2$\pm$6.6 & \textcolor{magenta}{\textbf{0.95}} & \textcolor{violet}{\textbf{0.94}} & \textcolor{orange}{\textbf{11}} & 7.3$\pm$2.1 & 20.2$\pm$6.7 \\ \hline
credit-a & 690 & 16 & \textcolor{magenta}{\textbf{0.89}} & \textcolor{violet}{\textbf{0.89}} & 944 & 10.1$\pm$2.6 & 21.4$\pm$5.6 & 0.83 & 0.83 & 84 & 10.3$\pm$3.8 & 23.3$\pm$9.8 & 0.85 & 0.85 & \textcolor{orange}{\textbf{36}} & \textcolor{blue}{\textbf{2.4$\pm$1.2}} & \textcolor{red}{\textbf{5.8$\pm$5.5}} \\ \hline
breast-w & 699 & 10 & 0.93 & 0.90 & 319 & 14.4$\pm$1.6 & 19.9$\pm$3.4 & \textcolor{magenta}{\textbf{0.95}} & \textcolor{violet}{\textbf{0.96}} & 34 & 10.5$\pm$1.9 & 18.6$\pm$5.4 & 0.94 & 0.92 & \textcolor{orange}{\textbf{9}} & \textcolor{blue}{\textbf{3.5$\pm$0.5}} & \textcolor{red}{\textbf{6.3$\pm$1.4}} \\ \hline
autism & 704 & 18 & \textcolor{magenta}{\textbf{0.93}} & \textcolor{violet}{\textbf{0.95}} & 359 & 10.3$\pm$2.0 & 25.2$\pm$5.8 & \textcolor{magenta}{\textbf{0.93}} & \textcolor{violet}{\textbf{0.95}} & 62 & 25.4$\pm$1.8 & 54.8$\pm$5.7 & 0.91 & 0.94 & \textcolor{orange}{\textbf{29}} & \textcolor{blue}{\textbf{9.9$\pm$1.3}} & \textcolor{red}{\textbf{23.6$\pm$4.3}} \\ \hline
parkinson & 765 & 754 & 0.70 & 0.78 & 159,556 & 8.9$\pm$1.8 & 13.4$\pm$3.0 & \textcolor{magenta}{\textbf{0.82}} & \textcolor{violet}{\textbf{0.89}} & 10,757 & 13.7$\pm$4.2 & 21.2$\pm$7.5 & \textcolor{magenta}{\textbf{0.82}} & \textcolor{violet}{\textbf{0.89}} & \textcolor{orange}{\textbf{9,691}} & \textcolor{blue}{\textbf{5.7$\pm$1.2}} & \textcolor{red}{\textbf{12.5$\pm$3.9}} \\ \hline
diabetes & 768 & 9 & 0.58 & 0.56 & 511 & 8.7$\pm$0.5 & 14.8$\pm$1.8 & 0.74 & 0.80 & 66 & 8.3$\pm$3.2 & 19.4$\pm$8.8 & \textcolor{magenta}{\textbf{0.75}} & \textcolor{violet}{\textbf{0.81}} & \textcolor{orange}{\textbf{38}} & \textcolor{blue}{\textbf{2.7$\pm$1.3}} & \textcolor{red}{\textbf{5.9$\pm$3.6}} \\ \hline
cars & 1728 & 7 & \textcolor{magenta}{\textbf{0.99}} & \textcolor{violet}{\textbf{0.99}} & 385 & 14.2$\pm$1.2 & 39.8$\pm$5.4 & 0.96 & 0.97 & 31 & 12.3$\pm$2.6 & 29.8$\pm$7.0 & 0.96 & 0.97 & \textcolor{orange}{\textbf{20}} & \textcolor{blue}{\textbf{7.2$\pm$1.4}} & \textcolor{red}{\textbf{14.0$\pm$3.2}} \\ \hline
kr vs. kp & 3196 & 37 & \textcolor{magenta}{\textbf{0.99}} & \textcolor{violet}{\textbf{0.99}} & 609 & 8.1$\pm$0.3 & 16.2$\pm$1.7 & \textcolor{magenta}{\textbf{0.99}} & \textcolor{violet}{\textbf{0.99}} & 226 & 19.3$\pm$3.8 & 46.7$\pm$11.0 & 0.97 & 0.97 & \textcolor{orange}{\textbf{152}} & \textcolor{blue}{\textbf{5.0$\pm$0.6}} & \textcolor{red}{\textbf{10.4$\pm$2.5}} \\ \hline
mushroom & 8124 & 23 & 1.0 & 1.0 & 923 & 8.3$\pm$0.8 & 12.7$\pm$1.2 & 1.0 & 1.0 & 281 & 7.9$\pm$0.3 & 11.9$\pm$0.3 & 1.0 & 1.0 & \textcolor{orange}{\textbf{254}} & \textcolor{blue}{\textbf{5.7$\pm$0.5}} & \textcolor{red}{\textbf{10.6$\pm$1.3}} \\ \hline
churn-model & 10000 & 11 & 0.54 & 0.60 & 9,941 & 11.6$\pm$2.5 & 39.2$\pm$8.0 & \textcolor{magenta}{\textbf{0.85}} & \textcolor{violet}{\textbf{0.91}} & 987 & 28.1$\pm$5.2 & 66.9$\pm$11.9 & \textcolor{magenta}{\textbf{0.85}} & \textcolor{violet}{\textbf{0.91}} & \textcolor{orange}{\textbf{600}} & \textcolor{blue}{\textbf{2.9$\pm$0.3}} & \textcolor{red}{\textbf{9.1$\pm$1.6}} \\ \hline
intention & 12330 & 18 & 0.88 & 0.93 & 8,542 & 25.2$\pm$6.7 & 91.6$\pm$27.6 & \textcolor{magenta}{\textbf{0.90}} & \textcolor{violet}{\textbf{0.94}} & 1,085 & 8.4$\pm$5.5 & 23.0$\pm$15.8 & \textcolor{magenta}{\textbf{0.90}} & \textcolor{violet}{\textbf{0.94}} & \textcolor{orange}{\textbf{661}} & \textcolor{blue}{\textbf{2.0$\pm$0.0}} & \textcolor{red}{\textbf{5.1$\pm$0.3}} \\ \hline
eeg & 14980 & 15 & 0.55 & 0.36 & 12,996 & 43.4$\pm$14.5 & 134.7$\pm$45.4 & \textcolor{magenta}{\textbf{0.72}} & \textcolor{violet}{\textbf{0.74}} & 2,735 & 69.1$\pm$17.3 & 152.6$\pm$35.6 & 0.67 & 0.68 & \textcolor{orange}{\textbf{1,227}} & \textcolor{blue}{\textbf{5.1$\pm$1.3}} & \textcolor{red}{\textbf{12.1$\pm$4.5}} \\ \hline
credit card & 30000 & 24 & 0.76 & 0.84 & 49,940 & 36.5$\pm$7.1 & 150.7$\pm$33.2 & \textcolor{magenta}{\textbf{0.82}} & \textcolor{violet}{\textbf{0.89}} & 5,954 & 19.1$\pm$6.4 & 48.8$\pm$14.7 & \textcolor{magenta}{\textbf{0.82}} & \textcolor{violet}{\textbf{0.89}} & \textcolor{orange}{\textbf{3,513}} & \textcolor{blue}{\textbf{2.0$\pm$0.0}} & \textcolor{red}{\textbf{3.0$\pm$0.0}} \\ \hline
adult & 32561 & 15 & 0.71 & 0.77 & 63,480 & 41.4$\pm$7.4 & 168.4$\pm$36.4 & \textcolor{magenta}{\textbf{0.84}} & \textcolor{violet}{\textbf{0.90}} & 2,508 & 16.8$\pm$2.4 & 46.7$\pm$10.4 & \textcolor{magenta}{\textbf{0.84}} & \textcolor{violet}{\textbf{0.90}} & \textcolor{orange}{\textbf{1,746}} & \textcolor{blue}{\textbf{2.0$\pm$0.0}} & \textcolor{red}{\textbf{5.0$\pm$0.0}} \\ \hline
rain in aus & 145460 & 24 & 0.63 & 0.70 & 3118,025 & 180.1$\pm$24.6 & 776.4$\pm$106.3 & 0.79 & 0.86 & 26,203 & 48.2$\pm$10.6 & 115.8$\pm$39.1 & \textcolor{magenta}{\textbf{0.82}} & \textcolor{violet}{\textbf{0.89}} & \textcolor{orange}{\textbf{10,243}} & \textcolor{blue}{\textbf{2.5$\pm$0.7}} & \textcolor{red}{\textbf{6.1$\pm$1.6}} \\ \hline \hline

average & 13873 & 57 & 0.81 & 0.82 & 180,475 & 23.6$\pm$4.2 & 82.8$\pm$15.7 & \textcolor{magenta}{\textbf{0.88}} & \textcolor{violet}{\textbf{0.91}} & 2,704 & 18.1$\pm$4.1 & 40.4$\pm$10.7 & \textcolor{magenta}{\textbf{0.88}} & 0.90 & \textcolor{orange}{\textbf{1,493}} & \textcolor{blue}{\textbf{4.2$\pm$0.8}} & \textcolor{red}{\textbf{9.2$\pm$2.6}} \\ \hline 

\end{tabular}
\caption{Comparison of RIPPER, FOLD-R++, and FOLD-SE on various Datasets}
\label{tbl:ripper}
\end{table*}

\section{Experimental Results}

\begin{table*}[!h]
\centering
\setlength{\tabcolsep}{2pt}
\small
\begin{tabular}{|l|c|c||c|c|c|c|c||c|c|c|c|c||c|c|c|c|c|}
\hline
\multicolumn{3}{|c||}{Data Set} & \multicolumn{5}{c||}{XGBoost} & \multicolumn{5}{c||}{MLP} & \multicolumn{5}{c|}{FOLD-SE} \\ \hline
Name & Rows & Cols & Acc & Prec & Rec & F1 & T(ms) & Acc & Prec & Rec & F1 & T(ms) & Acc & Prec & Rec & F1 & T(ms) \\ \hline
acute & 120 & 7 & \textcolor{magenta}{\textbf{1.0}} & 1.0 & \textcolor{orange}{\textbf{1.0}} & \textcolor{blue}{\textbf{1.0}} & 122 & 0.99 & 1.0 & 0.99 & 0.99 & 22 & \textcolor{magenta}{\textbf{1.0}} & 1.0 & \textcolor{orange}{\textbf{1.0}} & \textcolor{blue}{\textbf{1.0}} & \textcolor{red}{\textbf{1}} \\ \hline
heart & 270 & 14 & \textcolor{magenta}{\textbf{0.82}} & \textcolor{violet}{\textbf{0.83}} & \textcolor{orange}{\textbf{0.85}} & \textcolor{blue}{\textbf{0.83}} & 247 & 0.76 & 0.79 & 0.79 & 0.78 & 95 & 0.74 & 0.77 & 0.78 & 0.77 & \textcolor{red}{\textbf{13}} \\ \hline
ionosphere & 351 & 35 & 0.88 & 0.87 & 0.95 & 0.91 & 2,206 & 0.79 & \textcolor{violet}{\textbf{0.91}} & 0.74 & 0.81 & 1,771 & \textcolor{magenta}{\textbf{0.91}} & 0.89 & \textcolor{orange}{\textbf{0.98}} & \textcolor{blue}{\textbf{0.93}} & \textcolor{red}{\textbf{119}} \\ \hline
voting & 435 & 17 & 0.95 & \textcolor{violet}{\textbf{0.93}} & 0.95 & 0.93 & 149 & 0.95 & 0.92 & 0.94 & 0.93 & 43 & 0.95 & 0.92 & \textcolor{orange}{\textbf{0.96}} & \textcolor{blue}{\textbf{0.94}} & \textcolor{red}{\textbf{11}} \\ \hline
credit-a & 690 & 16 & \textcolor{magenta}{\textbf{0.85}} & 0.86 & \textcolor{orange}{\textbf{0.86}} & \textcolor{blue}{\textbf{0.86}} & 720 & 0.82 & 0.84 & 0.84 & 0.84 & 356 & \textcolor{magenta}{\textbf{0.85}} & \textcolor{violet}{\textbf{0.92}} & 0.79 & 0.85 & \textcolor{red}{\textbf{36}} \\ \hline
breast-w & 699 & 10 & 0.95 & 0.96 & \textcolor{orange}{\textbf{0.98}} & 0.96 & 186 & \textcolor{magenta}{\textbf{0.97}} & \textcolor{violet}{\textbf{0.98}} & 0.97 & \textcolor{blue}{\textbf{0.98}} & 48 & 0.94 & 0.88 & 0.97 & 0.92 & \textcolor{red}{\textbf{9}} \\ \hline
autism & 704 & 18 & \textcolor{magenta}{\textbf{0.97}} & 0.98 & \textcolor{orange}{\textbf{0.98}} & \textcolor{blue}{\textbf{0.98}} & 236 & 0.96 & \textcolor{violet}{\textbf{0.99}} & 0.96 & 0.97 & 56 & 0.91 & 0.94 & 0.94 & 0.94 & \textcolor{red}{\textbf{29}} \\ \hline
parkinson & 765 & 754 & 0.76 & 0.79 & 0.93 & 0.85 & 270,336 & 0.60 & 0.77 & 0.67 & 0.71 & 152,056 & \textcolor{magenta}{\textbf{0.82}} & \textcolor{violet}{\textbf{0.82}} & \textcolor{orange}{\textbf{0.96}} & \textcolor{blue}{\textbf{0.89}} & \textcolor{red}{\textbf{9,691}} \\ \hline
diabetes & 768 & 9 & 0.66 & 0.71 & 0.81 & 0.76 & 839 & 0.66 & 0.73 & 0.74 & 0.73 & 368 & \textcolor{magenta}{\textbf{0.75}} & \textcolor{violet}{\textbf{0.78}} & \textcolor{orange}{\textbf{0.85}} & \textcolor{blue}{\textbf{0.81}} & \textcolor{red}{\textbf{38}} \\ \hline
cars & 1728 & 7 & \textcolor{magenta}{\textbf{1.0}} & 1.0 & \textcolor{orange}{\textbf{1.0}} & \textcolor{blue}{\textbf{1.0}} & 210 & 0.99 & 1.0 & \textcolor{orange}{\textbf{1.0}} & \textcolor{blue}{\textbf{1.0}} & 83 & 0.96 & 1.0 & 0.94 & 0.97 & \textcolor{red}{\textbf{20}} \\ \hline
kr vs. kp & 3196 & 37 & \textcolor{magenta}{\textbf{0.99}} & \textcolor{violet}{\textbf{0.99}} & \textcolor{orange}{\textbf{1.0}} & \textcolor{blue}{\textbf{0.99}} & 403 &\textcolor{magenta}{\textbf{0.99}} & \textcolor{violet}{\textbf{0.99}} & \textcolor{orange}{\textbf{1.0}} & \textcolor{blue}{\textbf{0.99}} & 273 & 0.97 & 0.96 & 0.97 & 0.97 & \textcolor{red}{\textbf{152}} \\ \hline
mushroom & 8124 & 23 & 1.0 & 1.0 & \textcolor{orange}{\textbf{1.0}} & 1.0 & 697 & 1.0 & 1.0 & \textcolor{orange}{\textbf{1.0}} & 1.0 & 394 & 1.0 & 1.0 & 0.99 & 1.0 & \textcolor{red}{\textbf{254}} \\ \hline
churn-model & 10000 & 11 & \textcolor{magenta}{\textbf{0.85}} & 0.87 & \textcolor{orange}{\textbf{0.96}} & \textcolor{blue}{\textbf{0.91}} & 97,727 & 0.81 & \textcolor{violet}{\textbf{0.90}} & 0.86 & 0.88 & 18,084 & \textcolor{magenta}{\textbf{0.85}} & 0.87 & 0.95 & \textcolor{blue}{\textbf{0.91}} & \textcolor{red}{\textbf{600}} \\ \hline
intention & 12330 & 18 & \textcolor{magenta}{\textbf{0.90}} & 0.93 & \textcolor{orange}{\textbf{0.95}} & \textcolor{blue}{\textbf{0.94}} & 171,480 & 0.81 & 0.89 & 0.88 & 0.89 & 41,992 & \textcolor{magenta}{\textbf{0.90}} & \textcolor{violet}{\textbf{0.95}} & 0.93 & \textcolor{blue}{\textbf{0.94}} & \textcolor{red}{\textbf{661}} \\ \hline
eeg & 14980 & 15 & 0.64 & 0.64 & \textcolor{orange}{\textbf{0.81}} & \textcolor{blue}{\textbf{0.71}} & 46,472 & \textcolor{magenta}{\textbf{0.69}} & 0.72 & 0.71 & \textcolor{blue}{\textbf{0.71}} & 9,001 & 0.67 & \textcolor{violet}{\textbf{0.74}} & 0.63 & 0.68 & \textcolor{red}{\textbf{1,227}} \\ \hline
credit card & 30000 & 24 & NA & NA & NA & NA & NA & NA & NA & NA & NA & NA & \textcolor{magenta}{\textbf{0.82}} & \textcolor{violet}{\textbf{0.83}} & \textcolor{orange}{\textbf{0.96}} & \textcolor{blue}{\textbf{0.89}} & \textcolor{red}{\textbf{3,513}} \\ \hline
adult & 32561 & 15 & \textcolor{magenta}{\textbf{0.87}} & \textcolor{violet}{\textbf{0.89}} & \textcolor{orange}{\textbf{0.95}} & \textcolor{blue}{\textbf{0.92}} & 424,686 & 0.81 & 0.88 & 0.87 & 0.87 & 300,380 & 0.84 & 0.86 & \textcolor{orange}{\textbf{0.95}} & 0.90 & \textcolor{red}{\textbf{1,746}} \\ \hline
rain in aus & 145460 & 24 & \textcolor{magenta}{\textbf{0.84}} & 0.85 & \textcolor{orange}{\textbf{0.96}} & \textcolor{blue}{\textbf{0.90}} & 385,456 & 0.81 & \textcolor{violet}{\textbf{0.86}} & 0.89 & 0.88 & 243,990 & 0.82 & 0.85 & 0.94 & 0.89 & \textcolor{red}{\textbf{10,243}} \\ \hline \hline


average$^{*}$ & 12977 & 59 & \textcolor{magenta}{\textbf{0.88}} & 0.89 & \textcolor{orange}{\textbf{0.94}} & \textcolor{blue}{\textbf{0.91}} & 77,914 & 0.86 & \textcolor{violet}{\textbf{0.90}} & 0.88 & 0.89 & 42,735 & \textcolor{magenta}{\textbf{0.88}} & \textcolor{violet}{\textbf{0.90}} & 0.92 & 0.90 & \textcolor{red}{\textbf{1,381}} \\ \hline


\end{tabular}
\caption{Comparison of XGBoost, MLP, and FOLD-SE on various Datasets \\
\footnotesize \normalfont credit card dataset is excluded in computing average$^{*}$}
\label{tbl:stoa}
\end{table*}


We next present our experiments on UCI benchmarks and Kaggle datasets. The XGBoost Classifier is a well-known classification machine learning method and used as a baseline model in our experiments. Multi-Layer Perceptron (MLP) is another widely-used method that is able to deal with generic classification tasks. The settings used for XGBoost and MLP models is kept simple without limiting their performance. However, Both XGBoost and MLP models cannot directly perform training on mixed-type (numerical and categorical values in a row or a column) data. For mixed-type data, one-hot encoding has been used for data preparation 
in both 
because label encoding would add non-existing numerical relation to categorical values. RIPPER system is another rule-induction algorithm that generates formulas in conjunctive normal form as an explainable model. FOLD-R++ is the foundation of our new FOLD-SE algorithm. Both RIPPER and FOLD-R++ are capable of dealing with mixed-type data and are used as baseline to compare explainability. For binary classification tasks, accuracy, precision, recall, and F$_1$ score have been used as evaluation metrics of classification performance. For multi-category classification, accuracy and weighted F$_1$ score have been reported. The number of generated rules and the number of generated literals (predicates) have been used as evaluation metrics of explainability. 

The FOLD-SE algorithm does not need any data encoding for training, a feature that it inherits from FOLD-R++. After specifying the numerical features, both FOLD-R++ and FOLD-SE can deal with mixed-type data directly. Even missing values are handled and do not need to be provided. FOLD-SE has been implemented in Python. 
The hyper-parameter $ratio$ of these two algorithms is simply set to default value of 0.5 for all experiments. The hyper-parameter $tail$ of the FOLD-SE algorithms is set to default percentage 0.5\% of training data size. All the training processes have been performed on a small form factor desktop with Intel i7-8705G and 16 GB RAM. To have good performance test, we performed 10-fold cross-validation test on each dataset. We report average classification metrics and execution times. 

\subsection{\Large FOLD-SE {\it vs} FOLD-R++ and RIPPER} 

The experimental results shown in Table \ref{tbl:ripper} indicate that the FOLD-R++ and FOLD-SE algorithms outperform RIPPER algorithm in accuracy and explainability (the numbers of generated rules and literals/predicates). The FOLD-SE algorithm outperforms FOLD-R++ in explainability while maintaining comparable performance in classification, especially for large datasets. Note that in table \ref{tbl:ripper}, among the three methods, the various comparison metrics for best performer are highlighted in different colors: accuracy in pink, F1-score in purple, number of rules in blue, execution time in orange, and total number of predicates in the generated rule set in red. Note that in all the tables, for rules and number of predicates, we report averages for 10-fold cross-validation, which results in fractional values. We report (avg $\pm$ s.d.) for both.

Even for large datasets, 
the FOLD-SE algorithm can generate really concise rules that can capture patterns in datasets. As an example, for the Rain-in-Australia Dataset, FOLD-SE generates a model with 2.5 rules on average with 6.1 literals in these rules on average with average accuracy of 0.82, while RIPPER and FOLD-R++ report much higher values for number of rules and literals (180.1 rules and 776.4 literals for RIPPER and 48.2 rules and 115.8 predicates for FOLD-R++) and lower value for accuracy (0.63 for RIPPER and 0.79 for FOLD-R++).

\subsection{Compring FOLD-SE to XGBoost and MLP}

The experimental results comparing FOLD-SE with XGBoost and MLP are listed in Table \ref{tbl:stoa}. FOLD-SE \textit{always} takes less time to train compared to XGBoost and MLP, especially for large mixed-type datasets with many unique values. FOLD-SE can achieve equivalent scores wrt accuracy and F$_1$ score. For the ``credit card'' dataset, XGBoost and MLP failed training because of 16 GB memory limitation of the testing machine, the one-hot encoding process needs around 39 GB memory consumption. However, FOLD-SE only consumes 53 MB memory at peak for training on the same dataset. Compared to XGBoost and MLP, the FOLD-SE algorithm is more efficient and light-weight.

\begin{table*}[!h]
\centering
\setlength{\tabcolsep}{2pt}
\small
\begin{tabular}{|l|c|c||c|c|c||c|c|c||c|c|c|c|c||c|c|c|c|c|}
\hline
\multicolumn{3}{|c||}{Data Set} & \multicolumn{3}{c||}{XGBoost} & \multicolumn{3}{c||}{MLP} & \multicolumn{5}{c||}{FOLD-RM} & \multicolumn{5}{c|}{FOLD-SE-M}\\ \hline
Name & Rows & Cols & Acc & F1 & T(ms) & Acc & F1 & T(ms) & Acc & F1 & T(ms) &Rules &Preds & Acc & F1 & T(ms) &Rules &Preds \\ \hline
wine & 178 & 14 & 0.97 & 0.97 & 157 & \textcolor{magenta}{\textbf{0.98}} & \textcolor{violet}{\textbf{0.98}} & \textcolor{orange}{\textbf{9}}  & 0.93 & 0.94 & 33 & 7.5$\pm$0.5 & \textcolor{red}{\textbf{7.5$\pm$0.5}}  & 0.95 & 0.95 & 21 & \textcolor{blue}{\textbf{6.5$\pm$0.5}} & 7.6$\pm$0.5\\ \hline
ecoli & 336 & 9 & 0.54 & 0.51 & 1,182 & 0.65 & 0.63 & 198  & 0.79 & \textcolor{violet}{\textbf{0.80}} & 82 & 43.9$\pm$2.2 & 60.0$\pm$4.0  & \textcolor{magenta}{\textbf{0.80}} & 0.79 & \textcolor{orange}{\textbf{49}} & \textcolor{blue}{\textbf{24.0$\pm$4.2}} & \textcolor{red}{\textbf{45.2$\pm$11.4}} \\ \hline
weight-lift & 4024 & 155 & \textcolor{magenta}{\textbf{1.0}} & \textcolor{violet}{\textbf{1.0}} & 398,029 & 0.92 & 0.91 & 20,921  & \textcolor{magenta}{\textbf{1.0}} & \textcolor{violet}{\textbf{1.0}} & 3,716 & 14.7$\pm$0.6 & 16.9$\pm$1.4  & \textcolor{magenta}{\textbf{1.0}} & \textcolor{violet}{\textbf{1.0}} & \textcolor{orange}{\textbf{1,916}} & \textcolor{blue}{\textbf{7.0$\pm$0}} & \textcolor{red}{\textbf{10.6$\pm$0.7}}\\ \hline
wall-robot & 5456 & 25 & \textcolor{magenta}{\textbf{1.0}} & \textcolor{violet}{\textbf{1.0}} & \textcolor{orange}{\textbf{821}} & 0.93 & 0.93 & 3,423  & 0.99 & 0.99 & 4,207 & 29.6$\pm$2.4 & 41.6$\pm$3.6  & 0.99 & 0.99 & 2,440 &\textcolor{blue}{\textbf{ 7.1$\pm$0.3}} & \textcolor{red}{\textbf{15.8$\pm$0.9}}\\ \hline
page-blocks & 5473 & 11 & \textcolor{magenta}{\textbf{0.97}} & \textcolor{violet}{\textbf{0.97}} & 924 & \textcolor{magenta}{\textbf{0.97}} & \textcolor{violet}{\textbf{0.97}} & 1,971  & \textcolor{magenta}{\textbf{0.97}} & \textcolor{violet}{\textbf{0.97}} & 1,999 & 74.8$\pm$11.4 & 161.9$\pm$37.7  & 0.96 & 0.96 & \textcolor{orange}{\textbf{377}} & \textcolor{blue}{\textbf{8.5$\pm$1.7}} & \textcolor{red}{\textbf{15.0$\pm$3.0}}\\ \hline
nursery & 12960 & 9 & \textcolor{magenta}{\textbf{1.0}} & \textcolor{violet}{\textbf{1.0}} & 1,885 & \textcolor{magenta}{\textbf{1.0}} & \textcolor{violet}{\textbf{1.0}} & 1,083  & 0.96 & 0.96 & 1,169 & 59.5$\pm$5.3 & 142.0$\pm$15.5  & 0.92 & 0.92 & \textcolor{orange}{\textbf{525}} & \textcolor{blue}{\textbf{18.4$\pm$1.8}} & \textcolor{red}{\textbf{39.9$\pm$4.5}}\\ \hline
dry-bean & 13611 & 17 & \textcolor{magenta}{\textbf{0.93}} & \textcolor{violet}{\textbf{0.93}} & 6,032 & \textcolor{magenta}{\textbf{0.93}} & \textcolor{violet}{\textbf{0.93}} & \textcolor{orange}{\textbf{5,445}}  & 0.91 & 0.91 & 21,371 & 189.4$\pm$19.1 & 353.7$\pm$40.5  & 0.90 & 0.90 & 9,465 & \textcolor{blue}{\textbf{15.1$\pm$2.3}} & \textcolor{red}{\textbf{31.4$\pm$3.4}}\\ \hline
shuttle & 58000 & 10 & 1.0 & \textcolor{violet}{\textbf{1.0}} & 3,057 & 1.0 & \textcolor{violet}{\textbf{1.0}} & 17,126  & 1.0 & \textcolor{violet}{\textbf{1.0}} & 5,727 & 27.2$\pm$1.5 & 35.4$\pm$2.6  & 1.0 & 0.99 & \textcolor{orange}{\textbf{1,238}} & \textcolor{blue}{\textbf{4.0$\pm$0}} & \textcolor{red}{\textbf{5.0$\pm$0}}\\ \hline \hline 

average & 12505 & 31 & 0.93 & 0.92 & 51,513 & 0.92 & 0.92 & 6,272  & \textcolor{magenta}{\textbf{0.95}} & \textcolor{violet}{\textbf{0.95}} & 4,788 & 55.8$\pm$5.4 & 102.4$\pm$13.2  & 0.94 & 0.94 & \textcolor{orange}{\textbf{2,004}} & \textcolor{blue}{\textbf{11.3$\pm$1.4}} & \textcolor{red}{\textbf{21.3$\pm$3.1}}\\ \hline

\end{tabular}
\caption{Comparison of XGBoost, MLP, FOLD-RM, and FOLD-SE-M}
\label{tbl:foldsem}
\end{table*}

\subsection{FOLD-SE {\it vs} Decision Tree}

Table \ref{tbl:treefoldse} shows the results of comparison between decision tree and FOLD-SE. Nodes of a binary decision tree are equivalent to predicates of FOLD-SE. 
This is because both compare a data feature with a trained value then evaluate it as positive or negative. Therefore, the number of decision-tree nodes and the number of literals (predicates) in FOLD-SE rule-set, intuitively, are comparable. The rule sets generated by FOLD-SE is much easier to comprehend than the generated binary decision trees because there are far fewer predicates in the generated rule sets. For example, the generated binary decision tree for UCI Adult Dataset consists of over 4,000 nodes while the logic program generated by FOLD-SE only contains 5 predicates. As Table \ref{tbl:treefoldse} shows, FOLD-SE is also considerably faster in terms of execution time.

In summary, the rule-set generated by FOLD-SE is considerably more succinct than the Decision Tree method, RIPPER, and FOLD-R++. The execution time taken by FOLD-SE to generate the rule-set is also significantly lower.

\subsection{FOLD-SE for Multi-class Classification}

The FOLD-RM algorithm is the FOLD-R++ algorithm with the ``M'' extension for multi-class classification tasks. The ``M'' extension can also work with the FOLD-SE algorithm, we only need to add another parameter $tail$. The extended FOLD-SE is called FOLD-SE-M. Table \ref{tbl:foldsem} compares only FOLD-RM and FOLD-SE-M algorithms (we didn't find open source RIPPER implementation for multi-class classification). The hyper-parameter is also set as 0.5\% of training data size. \textit{The FOLD-SE algorithm keeps the resulting rule set smaller.}

\begin{table}[]
\centering
{
\small
\setlength{\tabcolsep}{2pt}
\begin{tabular}{|l|c|c||c|c|c||c|c|c|}
\cline{1-9}
\multicolumn{3}{|c||}{Data Set} & \multicolumn{3}{c||}{Decision Tree} & \multicolumn{3}{c|}{FOLD-SE} \\
\cline{1-9}
Name & Rows & Cols & Acc & Nodes & T(ms) & Acc & Preds & T(ms) \\
\cline{1-9}
acute & 120 & 7    & 1.0 & 4.0$\pm$0.0 & 3    & 1.0 & \textcolor{blue}{\textbf{3.0$\pm$0.0}} & \textcolor{red}{\textbf{1}} \\
\cline{1-9}
heart & 270 & 14    & \textcolor{orange}{\textbf{0.75}} & 38.8$\pm$3.4 & 38    & 0.74 & \textcolor{blue}{\textbf{9.1$\pm$8.5}} & \textcolor{red}{\textbf{13}} \\
\cline{1-9}
ionosphere & 351 & 35    & 0.88 & 19.0$\pm$1.0 & 377    & \textcolor{orange}{\textbf{0.91}} & \textcolor{blue}{\textbf{7.1$\pm$0.7}} & \textcolor{red}{\textbf{119}} \\
\cline{1-9}
kidney & 400 & 25    & 0.98 & 8.0$\pm$0.8 & 53    & \textcolor{orange}{\textbf{1.0}} & \textcolor{blue}{\textbf{6.1$\pm$0.3}} & \textcolor{red}{\textbf{16}} \\
\cline{1-9}
voting & 435 & 17    & 0.95 & 23.9$\pm$3.3 & 27    & 0.95 & \textcolor{blue}{\textbf{20.2$\pm$6.7}} & \textcolor{red}{\textbf{11}} \\
\cline{1-9}
credit-a & 690 & 16    & 0.80 & 75.4$\pm$3.1 & 137    & \textcolor{orange}{\textbf{0.85}} & \textcolor{blue}{\textbf{5.8$\pm$5.5}} & \textcolor{red}{\textbf{36}} \\
\cline{1-9}
breast-w & 699 & 10    & 0.92 & 35.3$\pm$1.4 & 48    & \textcolor{orange}{\textbf{0.94}} & \textcolor{blue}{\textbf{6.3$\pm$1.4}} & \textcolor{red}{\textbf{9}} \\
\cline{1-9}
autism & 704 & 18    & \textcolor{orange}{\textbf{0.92}} & 46.1$\pm$3.0 & 48    & 0.91 & \textcolor{blue}{\textbf{23.6$\pm$4.3}} & \textcolor{red}{\textbf{29}} \\
\cline{1-9}
parkinson & 756 & 764    & 0.81 & 34.9$\pm$1.9 & 18,253    & \textcolor{orange}{\textbf{0.82}} & \textcolor{blue}{\textbf{12.5$\pm$3.9}} & \textcolor{red}{\textbf{9,691}} \\
\cline{1-9}
diabetes & 768 & 9    & 0.68 & 117.4$\pm$5.4 & 166    & \textcolor{orange}{\textbf{0.75}} & \textcolor{blue}{\textbf{5.9$\pm$3.7}} & \textcolor{red}{\textbf{38}} \\
\cline{1-9}
cars & 1728 & 7    & \textcolor{orange}{\textbf{1.0}} & 47.9$\pm$1.2 & 53    & 0.96 & \textcolor{blue}{\textbf{14.0$\pm$3.2}} & \textcolor{red}{\textbf{20}} \\
\cline{1-9}
kr vs. kp & 3196 & 37    & \textcolor{orange}{\textbf{1.0}} & 43.9$\pm$1.1 & 422    & 0.97 & \textcolor{blue}{\textbf{10.4$\pm$2.5}} & \textcolor{red}{\textbf{152}} \\
\cline{1-9}
mushroom & 8124 & 23    & 1.0 & 11.9$\pm$0.3 & 1,463    & 1.0 & \textcolor{blue}{\textbf{10.6$\pm$1.3}} & \textcolor{red}{\textbf{254}} \\
\cline{1-9}
churn-model & 10000 & 11    & 0.80 & 1225.6$\pm$19.1 & 5,610    & \textcolor{orange}{\textbf{0.85}} & \textcolor{blue}{\textbf{9.1$\pm$1.6}} & \textcolor{red}{\textbf{600}} \\
\cline{1-9}
intention & 12330 & 18    & 0.86 & 843.2$\pm$13.2 & 6,886    & \textcolor{orange}{\textbf{0.90}} & \textcolor{blue}{\textbf{5.1$\pm$0.3}} & \textcolor{red}{\textbf{661}} \\
\cline{1-9}
eeg & 14980 & 15    & \textcolor{orange}{\textbf{0.85}} & 1141.9$\pm$16.4 &11,820    & 0.67 & \textcolor{blue}{\textbf{12.1$\pm$4.5}} & \textcolor{red}{\textbf{1,227}} \\
\cline{1-9}
credit card & 30000 & 24    & 0.73 & 3895.5$\pm$34.8 & 62,112    & \textcolor{orange}{\textbf{0.82}} & \textcolor{blue}{\textbf{3.0$\pm$0.0}} & \textcolor{red}{\textbf{3,513}} \\
\cline{1-9}
adult & 32561 & 16    & 0.82 & 4064.6$\pm$37.9 & 40,943    & \textcolor{orange}{\textbf{0.84}} & \textcolor{blue}{\textbf{5.0$\pm$0.0}} & \textcolor{red}{\textbf{1,746}} \\ 
\hline \hline
average & 6562 & 59    & 0.88 & 648.7$\pm$8.2 & 8,248    & 0.88 & \textcolor{blue}{\textbf{9.4$\pm$2.7}} & \textcolor{red}{\textbf{1,007}} \\
\cline{1-9}
\end{tabular}}
\caption{Comparison of Decision Tree, and FOLD-SE}
\label{tbl:treefoldse}
\end{table}

\section{Explainability}

\subsection{Examples}

With the new MGI heuristic, extended comparison operator, and rule pruning, the FOLD-SE algorithm pushes interpretability and explainability to a higher level. For Example \ref{ex:adult1} (UCI Adult Dataset), it generates the following logic program with \textit{only two rules!}:
 
{
\small
\begin{verbatim}
(1) income(X,'<=50K') :- 
        not marital_status(X,'Married-civ-spouse'), 
        capital_gain(X,N1), N1=<6849.0.
(2) income(X,'<=50K') :- 
        marital_status(X,'Married-civ-spouse'), 
        capital_gain(X,N1), N1=<5013.0, 
        education_num(X,N2), N2=<12.0.
\end{verbatim}  
}

\noindent
The above rules achieve 0.85\% accuracy, 0.86\% precision, 0.95\% recall, and 0.91\% F$_1$ score, the first rule covers 3457 test examples and the second rule covers 1998 test examples. 
The generated rule set can be understood easily due to the symbolic representation: Who makes less than \$50K dollars a year: (1) unmarried people with capital gain less than \$6,849; (2) married people with capital gain less than \$5,013 and education level not over 12. The generated rule-set for this dataset in the 10-fold cross validation test are almost all identical, only the values in the literals change slightly.

\begin{exmp}
\label{ex:parkinson1}
``Parkinson's Disease (PD)" is another classification task that contains 756 records with 754 features. We treat 80\% of the data as training examples and 20\% as testing examples. The task is to differentiate healthy people (class 1) from those with PD (class 0), the FOLD-SE algorithm generates the following rules: 
\end{exmp}

{
\small
\begin{verbatim}
(1) class(X,'1') :- std_6th_delta_delta(X,N1), N1=<0.027, 
        not ab3(X,'True').
(2) class(X,'1') :- std_6th_delta_delta(X,N1), 
        not(N1=<0.027).
(3) ab1(X,'True') :- locshimmer(X,N2), not(N2=<0.027), 
        tqwt_stdvalue_dec_11(X,N3), N3=<0.039, 
        tqwt_meanvalue_dec_3(X,N4), not(N4=<-0.0), 
        tqwt_kurtosisvalue_dec_27(X,N5), N5=<2.323.
(4) ab2(X,'True') :- tqwt_meanvalue_dec_3(X,N4), 
        not(N4=<-0.0), tqwt_tkeo_mean_dec_21(X,N6), 
        not(N6=<0.104), gq_std_cycle_open(X,N7), 
        N7=<33.937.
(5) ab3(X,'True') :- mean_mfcc_2nd_coef(X,N8), N8=<0.138,
        tqwt_kurtosisvalue_dec_36(X,N9), N9=<28.417, 
        not ab1(X,'True'), not ab2(X,'True').
\end{verbatim}  
}

\noindent
The generated rules achieve 0.83\% accuracy, 0.85\% precision, 0.94\% recall, and 0.89\% F$_1$ score.

\begin{exmp}
\label{ex:rain1}
``Rain in Australia" is another classification task that contains 145,460 records with 24 features. We treat 80\% of the data as training examples and 20\% as testing examples. The task is to find out if it is not rainy tomorrow, the FOLD-SE algorithm generates the following \textit{two rules!}: 
\end{exmp}

{
\small
\begin{verbatim}
(1) raintomorrow(X,'No') :- humidity3pm(X,N1), N1=<64.0, 
        rainfall(X,N2), N2=<182.6.
(2) raintomorrow(X,'No') :- rainfall(X,N2), N2=<2.2, 
        humidity3pm(X,N1), not(N1=<64.0), not(N1>81.0).
\end{verbatim}  
}

\noindent
The generated rules achieve 0.83\% accuracy, 0.85\% precision, 0.94\% recall, and 0.89\% F$_1$ score.

\subsection{Justification for Predictions}

FOLD-SE can give a justification for each prediction. Given a record, the information is turned into logic programming facts and the classification query executed on a logic programming systems. The proof tree generated serves as the justification for the query. FOLD-SE has a built-in justification-generation mechanism. Note that a rule set can also be translated into a natural language so that a non-expert can understand the model. Likewise, justification can also be provided in English. FOLD-SE comes equipped with these features. An example is shown below.

\begin{exmp}
\label{ex:titanic}
The ``Titanic Survival Prediction" is a classical classification challenge which contains 891 passengers as training examples and 418 passengers as testing examples and their survival based on features such as sex, age, number of siblings/spouses, number of parents/children, etc.. FOLD-SE generates the following program with 2 rules:
\end{exmp}

{
\small
\begin{verbatim}
(1) survived(X,'0') :- not sex(X,'female').
(2) survived(X,'0') :- class(X,'3'), sex(X,'female'), 
        fare(X,N1), not(N1=<23.25).
\end{verbatim}  
}

\noindent
The generated rules achieve 0.99\% accuracy, 0.98\% precision, 1.0\% recall, and 0.99\% F$_1$ score. Note that {\tt survived(X,0)} means that the person {\tt X} perished while {\tt survived(X,0)} does \textbf{not} hold means that the person person {\tt X} survived. Given a data sample {\tt Mr. James} with features: 

{
\small
\begin{verbatim}
    'number_of_siblings_spouses': 0.0, 'sex': 'male',
    'number_of_parents_children': 0.0, 'age': 34.5, 
    'fare': 7.8292, 'class': '3', 'embarked': 'Q'
\end{verbatim}
}

\noindent
The FOLD-SE makes the prediction that Mr. James perished with justification:
{
\small
\begin{verbatim}
(1) survived(Mr. James,'0') does hold because
        the value of sex is 'male' which should not 
        equal 'female' does hold 
\end{verbatim}
}

\noindent
Given another data sample {\tt Mrs. James} with features:
{
\small
\begin{verbatim}
    'number_of_siblings_spouses': 1.0, 'sex': 'female', 
    'number_of_parents_children': 0.0, 'age': 47.0, 
    'fare': 7.0, 'class': '3', 'embarked': 'S'
\end{verbatim}
}
\noindent
The FOLD-SE predicts that she survived with justification:
{
\small
\begin{verbatim}
(1) survived(Mrs. James,'0') does not hold because
        the value of sex is 'female' which should not 
            equal 'female' does not hold 
(2) survived(Mrs. James,'0') does not hold because
        the value of class is '3' which should equal 
            '3' does hold and
        the value of sex is 'female' which should equal 
            'female' does hold and
        the value of fare is 7.0 which should be 
            greater than 23.25 or be NaN does not hold
\end{verbatim}
}

\noindent One could argue that the English explanation is slightly clumsy, however, it can be easily improved (work is in progress).

\section{Related Work and Conclusions}


Rule-base Machine Learning is a long-standing interest of the research community. Some RBML algorithms perform training directly on the input data: ALEPH \cite{aleph} is a well-known Inductive Logic Programming algorithm that induces rules by using bottom-up approach, it cannot handle numerical features; those have to be handled manually. ILASP \cite{ilasp} system is another ILP algorithm that is able to generate normal logic program, it needs to work with a solver and requires a rule set to describe the hypothesis space. Some other RBML algorithms rely on statistical machine learning models: SVM+ProtoTypes \cite{svmplus} extracts rule from Support Vector Machine (SVM) models by using K-Means clustering algorithm. RuleFit \cite{rulefit} algorithm learns weighted rules from ensemble models of shallow decision trees. TREPAN \cite{trepan} produces a decision tree from trained Neural Networks by querying. Support Vector ILP \cite{svmilp} uses ILP-learned clauses as the kernel in dual form of SVM. nFOIL \cite{nfoil} system employs the naive Bayes criterion to guide its rule induction. The kFOIL \cite{kfoil} algorithm integrates the FOIL system with kernel methods. 
The TILDE \cite{tilde} algorithm generate propositional rules for paths from the root to every leaf node of a trained C4.5 decision tree. TILDE produces too many rules for large datasets when the generated decision tree has many leaf nodes.
Compared to the above systems, our approach is more efficient and scalable due to being top-down and using prefix-sum technique for literal selection. Thus, the rule-set learned in our approach is much more concise because of the use of default rules with exceptions and use of the Magic Genie Impurity heuristics. Finally, our approach is able to provide scalable explainability which, to the best of our knowledge, no other RBML algorithm achieves.

In this paper, we presented a highly efficient rule-based ML algorithm with scalable explainability, called FOLD-SE, to generate a normal logic program for classification tasks. FOLD-SE builds upon our previously developed FOLD-R++ algorithm by using newly developed heuristics for literal selection based on Gini Impurity and a rule pruning mechanism. Our experimental results show that the generated logic rule-set provides performance comparable to XGBoost and MLP but significantly better training efficiency, interpretability, and explainability. 
Unlike its predecessor FOLD-R++ as well as other RBML algorithms, the explainability of FOLD-SE is scalable which means the number of generated logic rules and generated literals stays small regardless of the size of the training data. The $tail$ hyper-parameter added to FOLD-SE can be easily adjusted to obtain a trade-off between accuracy and explainability. We also presented an extension of the FOLD-SE algorithm that can, similarly, handle multi-class classification tasks. 
 
\section*{Acknowledgement}

Authors acknowledge support from NSF grants IIS 1718945, IIS 1910131, IIP 1916206, US DoD. 

\bibliographystyle{ACM-Reference-Format}
\bibliography{mycitation}

\end{document}